\documentclass[table]{article}

\usepackage{PRIMEarxiv}

\usepackage[utf8]{inputenc} 
\usepackage[T1]{fontenc}    
\usepackage{hyperref}       
\usepackage{url}            
\usepackage{booktabs}       
\usepackage{amsthm, mathrsfs, amsmath,amssymb,amsfonts}       
\usepackage{nicefrac}       
\usepackage{microtype}      
\usepackage{lipsum}
\usepackage{fancyhdr}       
\usepackage{graphicx}       
\usepackage[table]{xcolor}
\definecolor{Low}{rgb}{1, 0.6, 0.6} 
\definecolor{LowMedium}{rgb}{1, 0.8, 0.6} 
\definecolor{Medium}{rgb}{1, 1, 0.6} 
\definecolor{MediumHigh}{rgb}{0.6, 1, 0.6} 
\definecolor{High}{rgb}{0.6, 1, 1} 
\usepackage{algorithm}%
\usepackage{algorithmicx}%
\usepackage{algpseudocode}%
\usepackage{listings}%
\usepackage{manyfoot}%
\usepackage{multirow}%
\graphicspath{{media/}}     
\usepackage{array}

\usepackage{pifont} 

\usepackage{subcaption}
\usepackage{tabularx}
\usepackage{booktabs}
\usepackage{url}
\usepackage{enumitem}
\usepackage[title]{appendix}
\title{A Multimodal Framework for Explainable Evaluation of Soft Skills in Educational Environments}

\author{
  Jared D.T. Guerrero-Sosa \\
  Department of Technologies and Information Systems \\
  University of Castilla-La Mancha, Spain\\
  \texttt{JaredDavid.Guerrero@uclm.es} \\
  \And
  Francisco P. Romero \\
  Department of Technologies and Information Systems \\
  University of Castilla-La Mancha, Spain \\
  \texttt{FranciscoP.Romero@uclm.es} \\
  \And
  Víctor Hugo Menéndez-Domínguez\\
  Mathematics School \\
  Autonomous University of Yucatan, Mexico \\
  \texttt{mdoming@correo.uady.mx} \\\And
  Jesus Serrano-Guerrero\\
  Department of Technologies and Information Systems \\
  University of Castilla-La Mancha, Spain \\
  \texttt{Jesus.Serrano@uclm.es} \\  \And
  Andres Montoro-Montarroso \\
  Department of Technologies and Information Systems \\
  University of Castilla-La Mancha, Spain \\
  \texttt{Andres.Montoro@uclm.es} \\
  \And
  Jose A. Olivas \\
  Department of Technologies and Information Systems \\
  University of Castilla-La Mancha, Spain \\
  \texttt{JoseAngel.Olivas@uclm.es} \\
}

\begin{document}
\maketitle

\begin{abstract}
In the rapidly evolving educational landscape, the unbiased assessment of soft skills is a significant challenge, particularly in higher education. This paper presents a fuzzy logic approach that employs a Granular Linguistic Model of Phenomena integrated with multimodal analysis to evaluate soft skills in undergraduate students. By leveraging computational perceptions, this approach enables a structured breakdown of complex soft skill expressions, capturing nuanced behaviours with high granularity and addressing their inherent uncertainties, thereby enhancing interpretability and reliability. Experiments were conducted with undergraduate students using a developed tool that assesses soft skills such as decision-making, communication, and creativity. This tool identifies and quantifies subtle aspects of human interaction, such as facial expressions and gesture recognition. The findings reveal that the framework effectively consolidates multiple data inputs to produce meaningful and consistent assessments of soft skills, showing that integrating multiple modalities into the evaluation process significantly improves the quality of soft skills scores, making the assessment work transparent and understandable to educational stakeholders.
\end{abstract}

\keywords{Multimodal Analysis \and Soft Skills Evaluation \and Fuzzy Logic \and Linguistic Perception Model}

\section{Introduction}
In recent years, the significance of soft skills, including decision-making, creativity, and communication, has been increasingly acknowledged in educational contexts. These skills are essential for personal and professional development~\cite{ putra2020}. Nevertheless, the accurate assessment of these skills remains a considerable challenge due to their intrinsic subjectivity, which is closely linked to human emotions and behaviour~\cite{Ulfa2024}. Given their complexity, soft skills require advanced assessment methods that can capture a range of expressive behaviours and adapt to individual differences among students.

The evaluation of soft skills can be classified according to whether the assessment is conducted offline or online \cite{Rasipuram2020}. Offline methods do not utilise digital technology, including public speaking, face-to-face interactions and group discussions. In contrast, online methods employ digital platforms to assess skills such as communication, collaboration and creativity in virtual environments. Some examples of online methods are video resumes and asynchronous video interviews (AVIs). 

A video resume is a digital tool that allows people to present themselves through a combination of verbal and non-verbal cues \cite{mestre2023}. In contrast, asynchronous video interviews entail more structured, prompt-driven responses, frequently employed in assessments or interviews \cite{Rasipuram2020}. These are scalable and can be conducted without the immediate presence of an interviewer. Although both methods are flexible, they are not without limitations in terms of depth. This is because they tend to focus on observable behaviours, which can obscure more subtle expressions of soft skills.

Traditional assessment techniques often fail to capture the subtle nuances of soft skills, especially in dynamic educational environments where students are exposed to a wide range of technological tools~\cite{Barthakur2023}, some based solely on machine learning or convolutional neural networks, resulting in limited assessments~\cite{Cheng2024}. These methods may be inadequate for the assessment of complex behavioural traits, which may lead to inconsistencies in the reliability and meaningfulness of soft skill evaluations.

Nowadays, the development of artificial intelligence (AI) technologies, coupled with advanced techniques such as fuzzy logic and deep learning, has enabled the interpretation of subtle human behaviours, including facial expressions \cite{Ngo2024}, emotions \cite{deMorais2024} and tone of voice \cite{Guerrero-Sosa2023}. These multimodal assessments integrate AI-driven analysis with linguistic perception models and can capture and quantify emotional and behavioural expressions, providing deeper insights into students' soft skills development~\cite{Feraco2023}. The utilisation of multimodal approaches facilitates the examination of a multitude of data types, collectively imparting a more intricate understanding of student interactions and engagement. Such comprehensive evaluations are necessary for establishing an educational environment that prepares students for the evolving demands of the workplace~\cite{BORGES2024107395}.

This paper introduces an approach which employs a Granular Linguistic Model of Phenomena (GLMP) integrated with multimodal analysis to assess soft skills in undergraduate students. The study applies the model to address the inherent uncertainties in emotional and behavioural expressions, thereby enhancing the interpretability and reliability of the assessments. This approach permits a systematic assessment of soft skills, whereby intricate behavioural indicators are analysed at various levels to yield meaningful insights. With this aim, this work addresses the following research questions (RQ): 

\begin{itemize}
\item RQ1: How can soft skills be objectively evaluated using multimodal analysis, specifically in educational environments?
\item RQ2: What are linguistic models and artificial intelligence roles in interpreting non-verbal and verbal cues in soft skills assessment?
\item RQ3: Can applying deep learning and fuzzy logic enhance the precision and interpretability of assessments in multimodal soft skills evaluation systems? 
\end{itemize}

The main contributions of this article are as follows:
\begin{itemize}
    \item Definition of a comprehensive framework for evaluating soft skills using video analysis, natural language processing, and emotion recognition techniques.
    \item Development of a multimodal assessment tool that utilises machine learning to analyse video, audio, and text data to assess various soft skills like decision-making, communication, and creativity.
    \item Advancement in the explainability of AI models used in education, providing more precise explanations of how decisions are made to assess soft skills.
\end{itemize}

This research article is organised as follows: Section \ref{sec:theoretical_framework} introduces the principal concepts and challenges associated with the evaluation of soft skills, covering multimodal analysis methods in the context of education and examines the GLMP. The methodology for multimodal soft skills analysis, including the assessment tool design and use of fuzzy logic and deep learning, is detailed in Section~\ref{sec:methods}. Section \ref{sec:results} focuses on evaluating decision-making, creativity, and communication skills among undergraduate students. Section \ref{sec:discussion_and_limitations} explores the implications of multimodal evaluation in education and the role of explainable AI in improving transparency. Finally, Section~\ref{sec:conclusions} summarises the study's contributions, limitations, and suggestions for future research.

\section{Theoretical Framework and Related Work}
\label{sec:theoretical_framework}

This section presents the fundamental concepts and challenges associated with the evaluation of soft skills. It also provides an overview of multimodal analysis approaches in educational environments and an examination of the Granular Linguistic Model of Phenomena for capturing complex behavioural assessments.

\subsection{Challenges in Soft Skills Evaluation}

The integration of soft skills into the technical undergraduate curriculum presents a number of challenges. Educators frequently encounter obstacles, including students' attitudes, limited time, and large class sizes, which complicate the teaching of soft skills in these programmes \cite{Idrus2018}. Notwithstanding these challenges, the development of soft skills is crucial for enhancing employability and performance in engineering and IT professions \cite{Zheng2015, BORGES2024107395, Lee2024}. In response to these challenges, some institutions have redesigned capstone courses to include team projects focused on soft skills training and assessment \cite{Zheng2015}. Others have introduced methodologies that incorporate soft skills without detracting from core technical content \cite{Cukierman2014}. Nevertheless, a considerable number of students do not fully recognise the relevance of soft skills to their future careers \cite{Naiem2015}. Research conducted in Egypt has identified significant deficiencies in students' understanding and proficiency in these skills, along with the necessity for universities to assume a more prominent role in promoting them \cite{Naiem2015}. These findings emphasise the sustained importance of effectively integrating soft skills into technical undergraduate education.

However, an emerging challenge is the use of large language models (LLMs) such as ChatGPT, which, while beneficial for generating text, code and prompt responses, can also make it difficult to assess students' comprehension, originality and critical thinking skills~\cite{Steele2023}. The advent of LLMs introduces several complexities in assessing skills~\cite{Huber2024} such as clarity, critical thinking and expression because these models may bypass traditional learning processes~\cite{Fleckenstein2024}. However, despite this, LLMs also offer opportunities for enhancing assessments by supporting linguistic perception models that can analyse both verbal and non-verbal cues \cite{Brin2023}. This aspect is directly related to the concept of explainability in the application of AI-based models, which involves obtaining easily interpretable models by humans, enabling domain experts to solve problems effectively~\cite{Ali2023,Vermesan2023}. Many educators view LLMs as tools that can be leveraged to develop new skills, such as verifying information, understanding context, and applying knowledge critically~\cite{Naamati2024, Moulin2024, deFine2023}.

\subsection{Multimodal Analysis in Educational Environments}

The utilisation of multimodal analysis in educational contexts has demonstrated favourable outcomes across a range of applications, facilitating enhanced learning experiences and assessment accuracy. A number of studies have highlighted the benefits of multimodal feedback for improving engagement and comprehension. For example, Hung \cite{Hung2016} emphasised the efficacy of multimodal video feedback in cultivating learner engagement, personalised learning, and active participation. 

In examining the integration of human and automated data sources, Worsley and Blikstein \cite{Worsley2018} illustrated that the combination of these multimodal data streams allows educators to gain detailed insights into student behaviours. Their research demonstrates that the combination of data sources reveals a multitude of patterns in learning, which is particularly valuable in understanding the intricate interactions that occur during hands-on activities. Similarly, Fjørtoft \cite{Fjortoft2020} examined the use of Multimodal Digital Classroom Assessments (MDCAs), finding that these assessments, when paired with traditional evaluation techniques, offer a comprehensive view of student learning processes and outcomes.

Du et al. \cite{Du2023} further expanded the use of diverse data types by incorporating electroencephalogram (EEG) data to assess cognitive load during collaborative learning. The results indicate that cognitive load is at its highest during the conceptualisation phase, offering valuable insights for task design and the development of tailored pedagogical strategies, particularly in collaborative settings. This study illustrates the potential of multimodal analysis, including video, to further refine collaborative learning methods.

In more advanced applications, Oh, Park, Lim, and Song \cite{Oh2024} introduced the Language Model Guided Matrix Factorization (LMgMF), which combines language models with various data types to predict student performance with greater accuracy. This approach addresses challenges such as the cold-start problem and enables the generation of personalised learning recommendations, which is particularly beneficial in contexts where data is limited. The capacity of LMgMF to function without altering the intrinsic mechanisms of language models evinces its adaptability and efficacy in educational contexts.

The field of embodied learning further expands the scope of multimodal analysis. Walkington, Nathan, Huang, Hunnicutt, and Washington \cite{Walkington2023} developed a framework for the analysis of students' interactions with physical and virtual objects, with a particular focus on gestures and movements. This approach employs augmented and virtual reality to enhance collaborative learning and provides insights into the design of educational technology by conceptualising cognition as a distributed process.

Petkovi\'c, Frenkel, Hellwich, and Lazarides \cite{Petkovic2025} put forth a computational model for evaluating non-verbal immediacy (NVI) in educational settings, integrating multimodal cues such as gesture intensity and perceived distance. By analysing a dataset of 400 labelled videos, this model effectively captures non-verbal behaviours and demonstrates a high degree of alignment with pedagogical outcomes when validated against human ratings. This underscores the significant impact of non-verbal communication on learning.

\subsection{Granular Linguistic Model of Phenomena}

Fuzzy logic enables the representation and processing of uncertainty and ambiguity, thereby accommodating the nuanced and often imprecise nature of human actions and responses, simulating the manner in which humans make decisions related to soft skills \cite{deNovais2024}. In order to adequately describe complex skills, it is necessary to employ a multi-level model. This approach is based on the concept of granularity, whereby data is organised into meaningful units, or granules, in order to capture the intricate details of a phenomenon within its unique context \cite{zadeh2002granular}.

In this structure, computational perceptions (CPs) provide a detailed perspective on specific elements within the system. In the context of modelling soft skills in situations such as soft skills evaluation, CPs provide insights based on subjective assessments of a range of different aspects.

A Granular Linguistic Model of Phenomena (GLMP) represents information granules within a PM network, with the level of detail adjusted as required. At each level, a CP can function as an input for the subsequent level, thereby enabling the incorporation of multiple layers of recursion. The Perception Mapping (PM) network establishes connections between input and output nodes and CPs via edges. The aggregation of CPs through PMs at higher levels of analysis produces new CPs, thereby enhancing the understanding and representation of the phenomenon. 

GLMP has been successfully applied across a range of contexts. For example, it has been employed to examine consumer consumption patterns and provide guidance for optimising consumption profiles \cite{TRIVINO201322}, generate automated linguistic reports on deforestation trends in the Amazon \cite{CONDECLEMENTE201746}, and develop a linguistic reporting system to describe process evolution based on real-time data and process patterns \cite{de2022natural}. Furthermore, these studies have contributed to the ongoing development and validation of the GLMP in a range of real-world applications. 

The components of the GLMP are described below:

A CP is defined as a tuple $(A, W, R)$, where:

\begin{itemize}
    \item The vector of linguistic expressions, represented by $A$, includes words or phrases in natural language and covers the entire linguistic scope of the CP. Each element, designated as $a_i$, is associated with the optimal linguistic description for a specific behavioural trait, taking into account a particular level of granularity. To illustrate, the creativity level could be described as $A = (low, medium, high)$.

    \item $W$: $(w_1, w_2, \dots, w_n)$ is a vector containing validity degrees, where $w_i \in [0, 1]$. The value of each $w_i$ indicates the degree of accuracy with which the linguistic expression $a_i$ represents the given input data, subject to the constraint that the sum of the $w_i$ values is equal to one.

    \item The relevance vector, denoted by $R$, comprises a set of values $(r_1, r_2, \dots, r_n)$, each of which represents the importance assigned to a given linguistic expression within a specific context. These values are assigned by the designer and may vary depending on the user type.
    
\end{itemize}

The utilisation of PMs serves to facilitate the processes of creation and combination of CPs. A PM integrates a plurality of input CP data points in order to generate a single, unified CP. A PM is represented as a tuple, denoted by $(U, y, g, T)$, where:

\begin{itemize}
\item The vector $U$, comprising $n$ input CP values $ (u_1, u_2, \dots, u_n)$, is represented as $u_i = (A_{ui}, W_{ui}, R_{ui})$. In the case of first-level PMs (1PM), the inputs may also include numerical values ($z$ $\in$ $R$) derived from measurement processes.

\item $y$ is the output CP, represented by the tuple $(A_y, W_y, R_y)$.

\item $g$ represents the aggregation function employed within the PM. In the field of fuzzy logic, a variety of functions are available for the nuanced handling of linguistic expressions. In the case of a first-level PM, the function $g$ typically employs membership functions in order to compute the degrees of validity and relevance.

\item $T$ is a text generation algorithm that produces sentences associated with the linguistic expressions in $A_y$. The function of $T$ is to serve as a linguistic template, encapsulating a range of possible expressions.

\end{itemize}

In educational settings, the use of GLMP offers distinct advantages over traditional modelling techniques. Unlike conventional black-box approaches, the hierarchical structure of GLMP enables a transparent and interpretable assessment process, which is relevant for educational stakeholders seeking to understand student performance beyond numerical scores. By representing behaviours and skills using linguistic labels grounded in expert knowledge, GLMP facilitates a more human-centred interpretation of learning outcomes. Moreover, its multi-level architecture allows for the integration of heterogeneous multimodal data, aligning with the inherently complex and context-dependent nature of soft skills. This makes GLMP particularly suitable for assessing nuanced competencies such as decision-making, communication, and creativity, where interpretability, adaptability, and contextualisation are essential. Consequently, GLMP not only enhances the explainability of AI-based evaluations but also supports personalised educational feedback and informed pedagogical decisions.

\section{Methods}
\label{sec:methods}

This study examines the efficacy of multimodal video analysis in evaluating fundamental soft skills, namely decision-making, communication, and creativity, among undergraduate students. The methodology employs a structured approach that combines expert-defined frameworks with data from multiple multimedia sources, thereby enabling a comprehensive assessment of student competencies. Subsequent subsections present an overview of the research target, design, and evaluation tool, outlining the processes employed to collect, process, and interpret data across educational environments.

\subsection{Research Target}

This case study examines the effectiveness of multimodal video analysis in assessing the soft skills of undergraduate students. Soft skills refer to abilities that cut across traditional academic disciplines and are relevant to various contexts, both within and outside of computing. These competencies are designed to prepare students for the multifaceted challenges they will face in their professional and personal lives by providing them with essential skills regardless of the specific computing discipline they are studying. The concept emphasizes the importance of developing well-rounded individuals who are technically proficient and capable of critical thinking, effective communication, ethical decision-making, and lifelong learning~\cite{Li2024}.

The soft skills considered in this work are the following:

\begin{itemize}
\item Decision-making: Act autonomously in learning, making informed decisions in different contexts~\cite{Beagon2023}.
\item Communication: Communicate effectively, both orally and in writing, adapting to the characteristics of the situation and the audience~\cite{Thornhill2023}.
\item Creativity: Propose creative and innovative solutions to complex situations or problems specific to the field of knowledge~\cite{Thornhill2023}.
\end{itemize}

A soft skill is constructed through a hierarchical assessment framework of dimensions, attributes and measures. Each soft skill has several dimensions that represent key aspects of the skill. Within each dimension, specific attributes are identified, and these attributes are quantified using measures derived from different multimedia modalities such as text, audio and video. This structured approach allows for a comprehensive and multi-faceted assessment of the student's soft skills. 

Following this structure, a GLMP was established for each soft skill, wherein the lowest level of granularity is represented by the measures obtained for the three multimedia modalities. The sequence of the PMs is as follows: Measures (First Order PMs), Attributes (Second Order PMs), Dimensions (Third Order PMs) and Soft Skill (Fourth/Top Order PMs). 

A panel of experts from the Autonomous University of Yucatan, comprising faculty from the Psychology and Education departments, conducted a definition of the assessable aspects of a student. The education experts identified the soft skills suitable for task assessment to complement traditional competencies with those required in the modern workplace. The dimensions, attributes, and measures for each soft skill were defined by psychology experts, who also established the scales and relevance of each aspect for evaluating students' skills. 

\begin{figure}[ht!]
    \centering
    \includegraphics[scale=0.07]{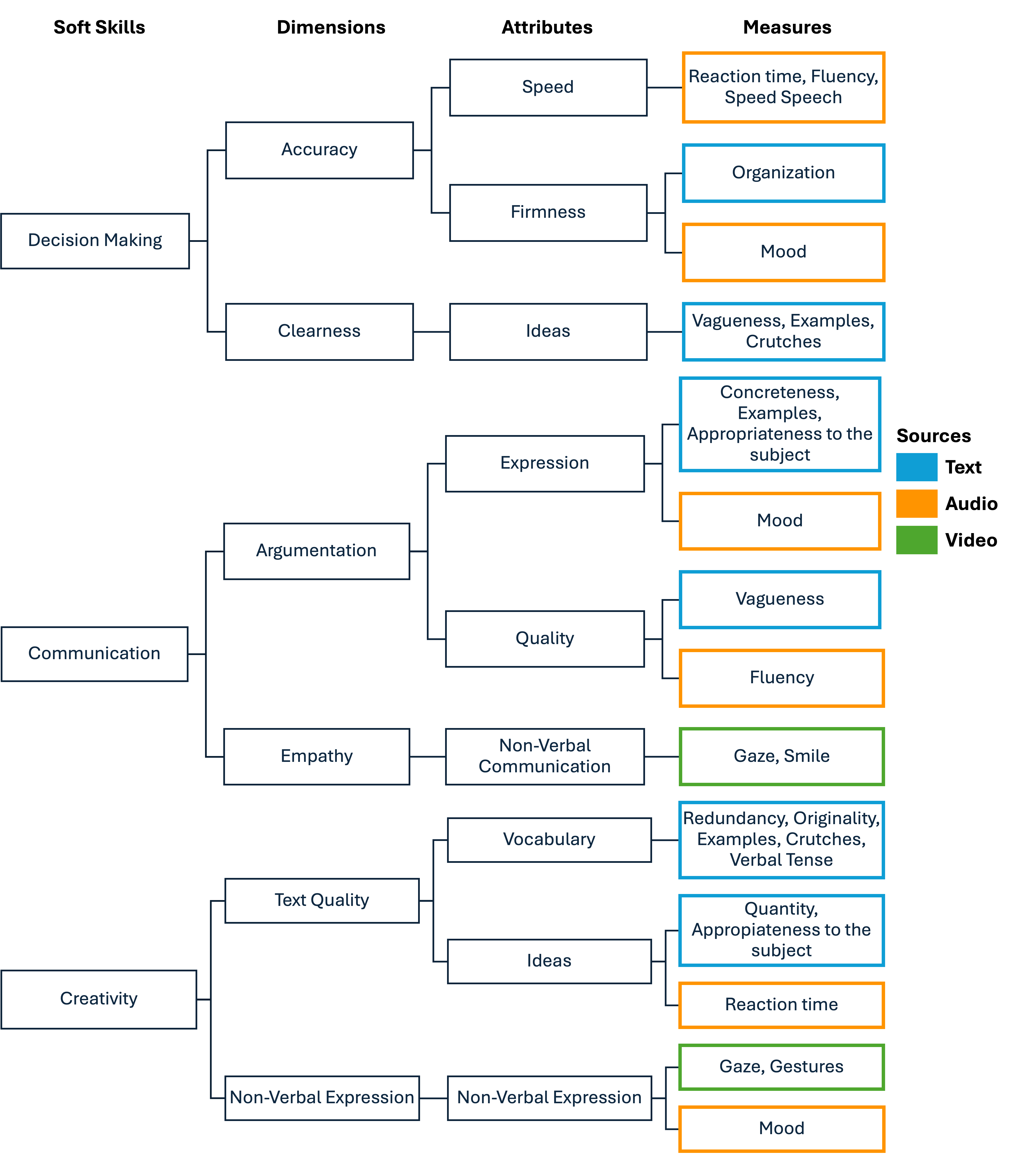}
    \caption{Dimensions, attributes and measures related to soft skills}
    \label{fig:concept}
\end{figure}

Furthermore, the panel underscored the significance of discerning micro-expressions, which are brief, involuntary facial expressions that manifest when an individual attempts to disguise their genuine emotions~\cite{Wang2020, Li2021}. These expressions, which can be challenging to discern with the naked eye, were identified as a relevant aspect in the assessment of soft skills. Additionally, the engineering team delineated the methodologies and techniques for extracting data from diverse multimedia sources and established the aggregation operations. Fig. \ref{fig:concept} describes the complete definition of each soft skill to evaluate in terms of dimensions, attributes and measures. 

Several steps, from video upload to soft skills assessment, are taken to ensure student privacy throughout the assessment process. Videos are uploaded via a secure, encrypted channel and immediately anonymised by replacing personal identifiers with unique codes. All data is stored in encrypted databases with strict access controls and is accessible only by authorised personnel. The assessment integrates different data modalities which are processed separately to ensure that no personal data is linked back to students. Results are presented in an aggregated, anonymised form without any personal identifiers. The system does not store student information; all multimedia data is deleted at the end of the assessment. Only the teacher can identify which assignment belongs to each student by keeping a secure, separate record of unique codes not shared with the research team.

\subsection{Research Design}

Two subjects from different universities (the University of Castilla La Mancha - UCLM - and the Autonomous University of Yucatan - UADY) have been used to verify the proposal's feasibility. The topics related to the subjects are closely linked to the tools used in the evaluation process, which is a factor in improving student participation. The experiment was conducted for each group throughout the course, with students completing three activities at different stages of the semester. The first activity took place at the beginning of the course, and a baseline for the participants' soft skills was established. The second activity was carried out in the middle of the course. By then, it was expected that students would have begun to develop their interpersonal and communication skills alongside their academic work. The final activity took place at the end of the semester and provided an opportunity to observe how these soft skills had developed over time. 

Regarding online evaluation methods, Task 1 can be classified as a video resume, where students present themselves through a pre-recorded video showcasing their communication and interpersonal skills. Tasks 2 and 3, on the other hand, align with the concept of asynchronous video interviews, in which students respond to structured questions without real-time interaction, enabling an assessment of their ability to articulate responses under more interview-like conditions. The purpose of this structure was to allow students sufficient time for the natural development of their soft skills while they continued their academic training. Each of the groups is described in detail below.

\subsubsection{Group 1: Machine Learning Course}

The Machine Learning course introduces the fundamental concepts, algorithms and applications of machine learning. It covers the theoretical foundations of machine learning algorithms and their practical implementation. Students will learn about supervised and unsupervised learning models, with a focus on data preparation and model evaluation. Machine Learning is one of the subjects of the Computer Engineering degree offered by the School of Computer Science of Ciudad Real of the University of Castilla La Mancha (Spain). The course is taught in English. The students are of different nationalities, mainly Spanish. 

Data from 28 college students were collected. Ten students comprise group A of students in the 2022 edition of the course, and the remaining 18 are from the 2023 edition. Student participants complete three activities (see Table \ref{tab_activities1}): a personal presentation, lecture summarization and topic report. Group A conducted task 1 through a video-recording and both task 2 and 3 through a writing report, on the contrary, group B carry out all activities through a video recording.

        \begin{table}[!ht]
            \small
            \caption{Machine Learning Course Activities}
            \label{tab_activities1}
            \centering
            \begin{tabular}{|l|p{8cm}| c | c |}
                \hline
                \multicolumn{2}{|c|}{List of Activities} & Group 22 & Group 23 \\
                 \hline
                   1 &      Give your data (name and surname), your specialization and computer science interests, and answer these questions: Why did you choose the Machine Learning subjects? What does the concept of “Machine learning” mean to you? & Video & Video\\
                   \hline
                   2 &
                   The task consists of creating a personalised summary of the key points and your opinions about a talk. The purpose is to assess the ability to synthesize and reflect on a topic. & Text & Video\\
                   \hline
                   3 &  The bonded task consists of creating a report about a topic focusing on the key points and future direction.   & Text & Video \\           
                  \hline
        \end{tabular}
\end{table}

\subsubsection{Group 2: Human-Computer Interaction Course}
Human-computer interaction is a compulsory subject in the Bachelor's degree programme in Software Engineering at the Autonomous University of Yucatan, taught at the Faculty of Mathematics in Mérida, Mexico. Sixth-semester students of 2024 were evaluated. The Human-Computer Interaction course addresses key concepts and aspects regarding human-computer interaction, such as user-centred design, interface analysis and design, as essential elements for contributing aspects to guarantee software product quality. Students are expected to develop software systems that provide adequate interactions for the user, considering methodologies and basic principles of interfaces in interactive systems. The subject is taught in Spanish, and all students are from Mexico.

Data from 21 university students was divided into two groups, the first with 10 students (Group A) and the second with 11 students (Group B). The formation of the two groups was done randomly. Group A comprised two women and eight men, whereas Group B comprised five women and six men. The average age in both groups was 20 years.
Participating students complete three activities (see table \ref{tab_activities2}): a personal presentation, a lesson summary and a final reflection on what they have learned in the course. All activities were conducted through video for group A, while for group B, the first and third tasks were conducted through video and the second by text.

        \begin{table}[!ht]
           \small
            \caption{Human-Computer Interaction Course Activities}
            \label{tab_activities2}
            \centering
            \begin{tabular}{|l|p{8cm}| c | c |}
                \hline
                \multicolumn{2}{|c|}{List of Activities} & Group A & Group B \\
                 \hline
                   1 &      Give your data (name and surname), your interests in Software Engineering and answer the following question: What do you think human-computer interaction is? & Video & Video\\
                   \hline
                   2 &
                   The task consists of creating a personalized summary of the key points and your opinions about a talk. . The purpose is to assess the ability to synthesize and reflect on a topic. The talk is from a virtual conference about ten design flaws that plague today's websites & Video & Text\\
                   \hline
                   3 &  Based on what you have studied in the course, reflect on what you have learnt in the course.   & Video & Video \\           
                  \hline
        \end{tabular}
\end{table}

\subsection {Evaluation Tool}

\begin{figure}[ht!]
    \centering
    \includegraphics[scale=0.3]{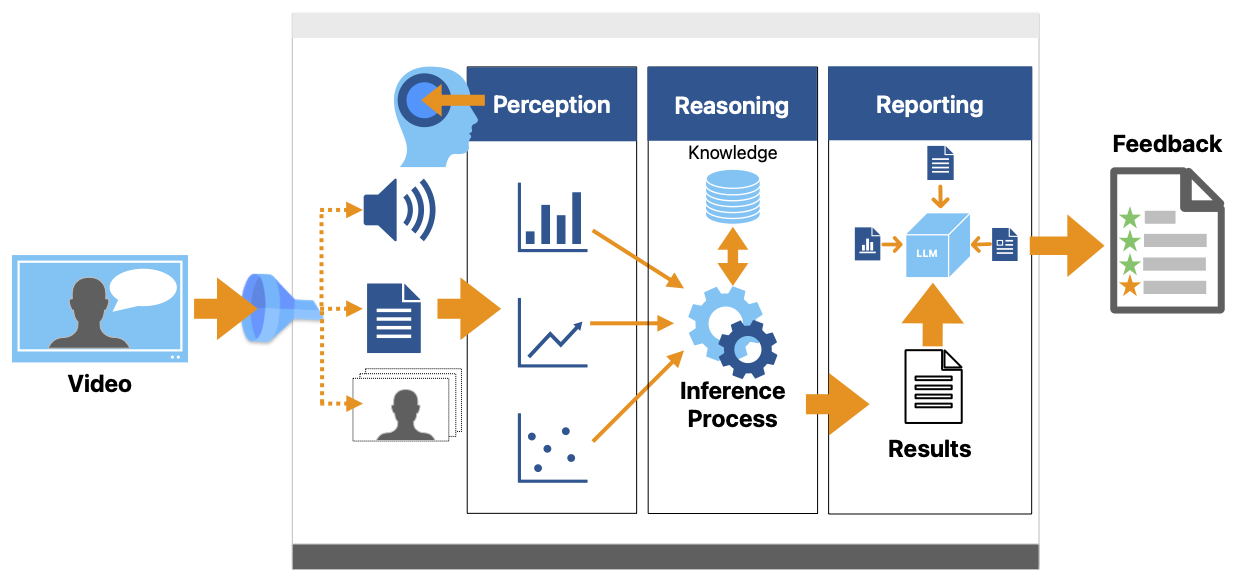}
    \caption{Video Evaluation Tool}
    \label{fig:video_evaluation_tool}
\end{figure}

The process of evaluating a student's competencies is conducted through a tool whose input is a video file (Fig. \ref{fig:video_evaluation_tool}). The tool is based on Deep Learning and Fuzzy Logic techniques. The pre-trained deep learning models allow for identifying and extracting human characteristics, which together describe human behaviours. On the other hand, the use of fuzzy logic facilitates the handling of uncertainty and possible imprecision due to aspects of the video that could obstruct the evaluation (e.g. lack of correct lighting, low-quality image), as well as the representation of each aspect of the evaluation with linguistic labels according to the context.

The incorporation of explainability into the evaluation process is facilitated by the output generated by the GLMP, as it provides interpretable and human-centred insights. By representing data through linguistic labels and leveraging fuzzy rule-based models, the tool ensures clarity in its decision-making. This explainability is crucial for establishing trust and mutual understanding among educators and stakeholders, as it translates intricate computational processes into outcomes that are intelligible and aligned with human reasoning \cite{Mencar2019}.

\subsubsection{Input Data}
\begin{figure}
    \begin{subfigure}{0.45\textwidth}
        \includegraphics[width=\linewidth]{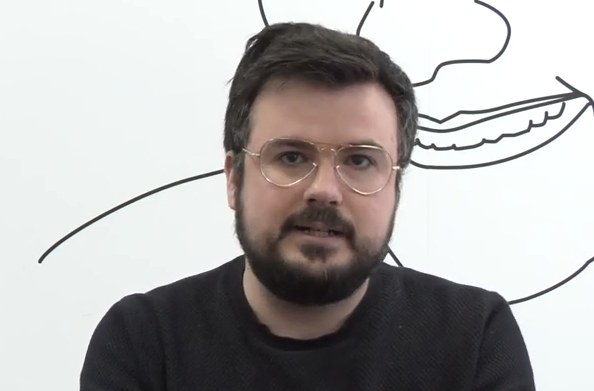}
        \caption{Minimum recommended distance} \label{fig:distance_a}
    \end{subfigure}%
    \hspace*{\fill}   
    \begin{subfigure}{0.45\textwidth}
        \includegraphics[width=\linewidth]{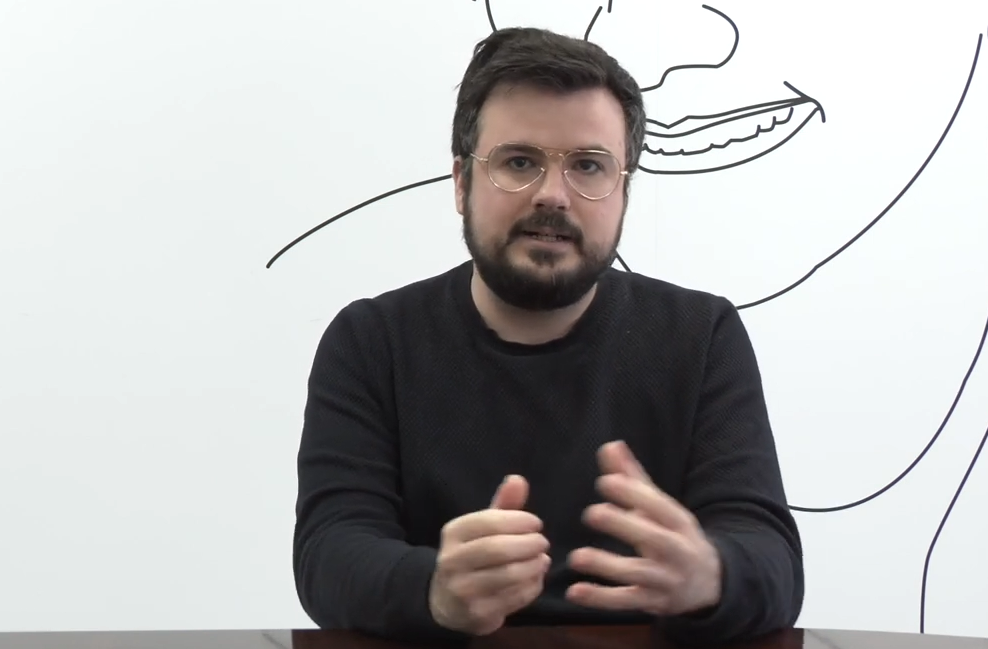}
        \caption{Maximum recommended distance} \label{fig:distance_b}
    \end{subfigure}%
\caption{Recommended distances for close-up video} \label{fig:distance}
\end{figure}

The input received by the tool is a valid video file. It is suggested that MP4 or MOV formats speed up the processes. If the video does not comply with the format, the input video is transformed into MP4 format. This video must have a resolution of 1280 x 720 in horizontal position. At all times, it must be possible to view the whole head of the person, looking in the direction of the camera. The minimum distance from the person to the camera must be sufficient to show the head and shoulders (Fig. \ref{fig:distance_a}), while the maximum distance allows the upper half of the person's body to be displayed (Fig. \ref{fig:distance_b}).

\subsubsection{Perception Process}
Machine learning and deep learning techniques allow for identifying the person's characteristics in the video, which are obtained according to the modality. Three multimedia modalities are extracted from the video: audio, text and image. 

\textbf{Audio:} The audio was extracted from the video and stored in a WAV file using the free software tool FFmpeg \cite{Newmarch2017}. Once the audio has been extracted, the person's prosody - the rhythm and tone used throughout a speech and how it relates to the message being conveyed - is analysed. This analysis is carried out by using the Python library MyProsody, whose analysis is based on vocal pitch, acoustic intensity and rhythm~\cite{Jadoul2018}.

\textbf{Text:} The text with the transcription of the speech is obtained employing Whisper as an automatic speech recognition system~\cite{Radford2023}. The transcription is used to analyse the content of the discourse. Discourse themes are obtained using a text classifier based on a list of categories defined by the controlled vocabulary taxonomy EuroVoc\footnote{\url{https://eur-lex.europa.eu/browse/eurovoc.html?locale=en}} which is used because of its broad categorization. On the other hand, dictionaries were defined in both languages to identify textual components in the transcription, such as terms of concreteness and argumentative effort, crutches, discourse connectors throughout the speech (beginning, middle and end) and common words in the language. Moreover, natural language processing (NLP) is employed to detect redundant terms and adjectives in text. This involves part-of-speech tagging using the Spacy library, enabling the retraining of existing models with customised data pertinent to specific domains~\cite{vasiliev2020}.

\textbf{Video:} The next step is to analyse some of the person's behaviour throughout the video to evaluate aspects of non-verbal language. To do this, the frames that make up the video are extracted and, using the OpenCV image processing software toolkit~\cite{Kumari2023}, the Haar cascade methodology is applied to identify the individual and facial attributes~\cite{Viola2001} such as eyes, face and smile. The dominant emotion of the person in each frame is also detected by DeepFace using its pre-trained model with the following classes: happiness, neutrality, fear, surprise, displeasure, sadness and anger~\cite{Taigman2014}. Gaze focus and blink detection are achieved using a dlib's 68-point facial landmark predictor. This pre-trained neural network identifies the 68 key points (also known as landmarks), defining the features of a human face in an image~\cite{Chandel2023}. This model detects whether a person looks left, right, centre, or blinking.

From the human characteristics extracted from the multimedia modalities, some measures are relevant for the subsequent identification of aspects related to the soft skills to be detected. Table \ref{tab_measures_first} lists the measures (previously shown in Fig. \ref{fig:concept}), with their name, their measurement method and the source where these methods are applied to obtain the corresponding values. For the handling of the different concepts from the basic measures, linguistic labels have been used, i.e. for each concept, three levels have been established (e.g. Low, Medium, High) that allowed a more flexible subsequent treatment, whatever the measurement scale that has been specifically defined for each measure.

\begin{table}[!htp]
    \caption{Measures Description}
    \label{tab_measures_first}
    \centering

    \begin{tabular}{|p{2cm}|p{8.5cm}|l|}
     \hline
     \textbf{Measure} & \textbf{Measurement method} & \textbf{Source} \\
     \hline
     Appropriateness to the subject & Importance within the text of the topics related to the subject matter of the questions. & Text \\ \hline
     Concreteness & Number of argumentative, reinforcing or concretizing linguistic markers per minute related to the number of topics covered. & Text \\ \hline
     Crutches & Number of phrases used in the text. & Text \\ \hline
     Examples & Number of adjectives per minute accompanying explanation. & Text \\ \hline
     Fluency & Number of pauses and interruptions. & Audio \\ \hline
     Gaze & Detection of whether the gaze is focused on a point or is furtive. The ratio of focused gazes over the total. & Video \\ \hline
     Gesture & Positive, negative and stress expressions are measured. & Video \\ \hline
     Mood & Detecting whether the person is reading, using a normal or passionate tone. & Audio \\ \hline
     Organization & Use of speech connectors (normalized per minute). & Text \\ \hline
     Originality & The use of specific and uncommon vocabulary is measured. Ratio of words used that are not frequently used. & Text\\ \hline
     Quantity & Number of topics addressed in the context. & Text \\ \hline
     Reaction time & Time it takes to decide or start talking (in seconds). & Audio \\ \hline
     Redundancy & Identification of word repetition. The number of times the most frequent words have been repeated (without stop words). & Text\\ \hline
     Speech speed & Syllables per second. & Audio \\ \hline
     Smile & Smile detection. & Video \\ \hline
     Vagueness & Ambiguous terms in the expression (per minute). & Text \\ \hline
     Verbal tense & Percentage of verbs expressed in the present tense out of the total number of verbs in the speech. & Text\\ \hline
    \end{tabular}
\end{table}

\subsubsection{Reasoning Process}
Through the acquisition of the knowledge of a panel of experts, it has been defined how each of the measures, attributes and dimensions that define soft skills at different levels lead to the creation of the knowledge base, which is built by a set of inference rules to be applied to draw conclusions. The procedure for evaluating each soft skill aggregates from the highest level of detail (soft skills) to the lowest level (measures).  Once the measures have been obtained, they are aggregated to obtain the dimensions. At each level of the model, the aggregation procedure depends on the number of inputs.

The aggregation uses fuzzy rules if the evaluated aspect consists of up to three inputs. The use of fuzzy rules in this context facilitates an accurate scoring process while maintaining computational simplicity and flexibility. When dealing with up to three measures, fuzzy rules allow for easier manipulation and interpretation, allowing the system to effectively incorporate expert knowledge. 

Otherwise, weighted average operators are used, where the relative importance of each measure is taken into account in defining the attribute, ensuring that more influential aspects contribute appropriately to the overall assessment.

The GLMP is employed in this reasoning process to provide a structured, hierarchical framework that facilitates a comprehensive assessment of soft skills through a granular approach. Each measure, attribute, and dimension is represented as a CP within the GLMP, which translates raw data into interpretable linguistic labels. This model permits each CP at one level to serve as an input for higher levels, thereby creating more comprehensive views of soft skills through a recursive process.

At the initial levels, the GLMP employs measures such as reaction time, fluency, and speech speed to define attributes with a high level of granularity. These attributes are subsequently aggregated into dimensions and soft skills, in accordance with expert-defined rules. By applying the GLMP's hierarchical structure, the model effectively represents soft skills across levels, thereby ensuring that the insights gained are both comprehensive and aligned with the nuances of human interaction.

The defined rules consist of a set of antecedents and a consequent, which can be an attribute, a dimension or a soft skill, as shown below:

\begin{center}
\noindent\fbox{%
    \parbox{0.95\textwidth}{%
    If \textbf{Reaction time} is \textit{Low} and \textbf{Fluency} is \textit{Medium} and \textbf{Speech speed} is \textit{High} then  \textbf{Speed} is \textit{Medium}. \\ \\
    If \textbf{Speed} is \textit{Low} and \textbf{Firmness} is \textit{High} then  \textbf{Accuracy} is \textit{Medium}. \\ \\
    If \textbf{Accuracy} is \textit{High} and \textbf{Clearness} is \textit{Medium} then  \textbf{Decision-making} is \textit{High}.

    }%
}
\end{center}

Once the rules have been evaluated, the pre-report is generated according to the hierarchical structure defined for each soft skill. Each level contains the corresponding component (soft skill, dimension, attribute or measure) and its output in a linguistic label to show the reasoning behind the results obtained transparently.  

\subsubsection{Reporting}
The results obtained in the reasoning phase are used as a template to generate a detailed report with LLM, for example LLaMA, which is a family of large language models designed for accurate and efficient text generation in a variety of tasks \cite{Touvron2023}. Therefore,  the results are presented concisely, emphasising low and high performers at the highest levels of detail and avoiding redundancy in cases where descriptions have the same linguistic label. The report is generated, providing a detailed assessment of each student's soft skills. This process guarantees that all student performance aspects are encompassed, providing an overview of their capabilities. The report incorporates each soft skill's specific dimensions, attributes, and measures, emphasising strengths and areas for improvement with contextual explanations. The format prevents technical jargon, ensuring the information is understandable for diverse educational stakeholders. This characteristic facilitates personalised feedback and targeted interventions to assist students in enhancing their soft skills. 

An example of a part of the pre-report and the final report for the decision-making soft skill is shown in Fig \ref{fig:report_example}. The following steps outline the report generation process:

\begin{figure}[ht!]
    \centering
    \includegraphics[scale=0.4]{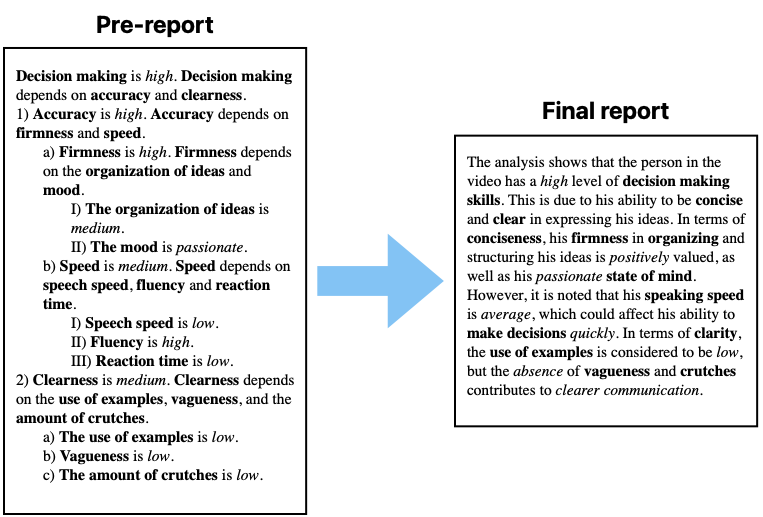}
    \caption{Example of a Soft Skill Report}
    \label{fig:report_example}
\end{figure}

\begin{enumerate}
    \item Data input: The pre-report, containing scores from the preliminary evaluation, is prepared and formatted as input for the model.
    \item Template application: LLM receives as input a pre-defined template to ensure consistency and structure in the final report. This template guides the model in organising the content, focusing on key aspects such as communication skills, decision-making and creativity. The model synthesises the input data, transforming detailed assessments into a narrative format. This process includes summarising overall performance, highlighting strengths and weaknesses and offering contextual explanations for scores, particularly in areas involving subjective interpretation (e.g., emotional expressions and non-verbal cues).
    \item Output: The final report is recorded, providing a coherent and accessible explanation of the output generated by the GLMP. This approach enhances the interpretability of the different levels of granularity within the model, making complex assessments of soft skills more understandable.
\end{enumerate}

Using an LLM model ensures that the language used in the report is academically appropriate, clear, and easily understandable. It avoids using technical terms wherever possible and explains any necessary terms, thus making the report accessible to a broad range of educational stakeholders. Prompt engineering techniques mitigate the potential for hallucination in the generated reports and guide the model's response~\cite{Bhattacharya2024}. 
Therefore, the prompts are augmented with contextual information from the report, including details on the skills, dimensions, attributes, and measures being assessed. Providing clear, structured, and contextually relevant prompts reduces the model's probability of generating erroneous or irrelevant information. Thus, it is guaranteed that the final report is reliable and aligned with the actual assessment data, thus enhancing the overall quality and trustworthiness of the output.

\begin{figure}[ht!]
    \centering
    \includegraphics[scale=0.14]{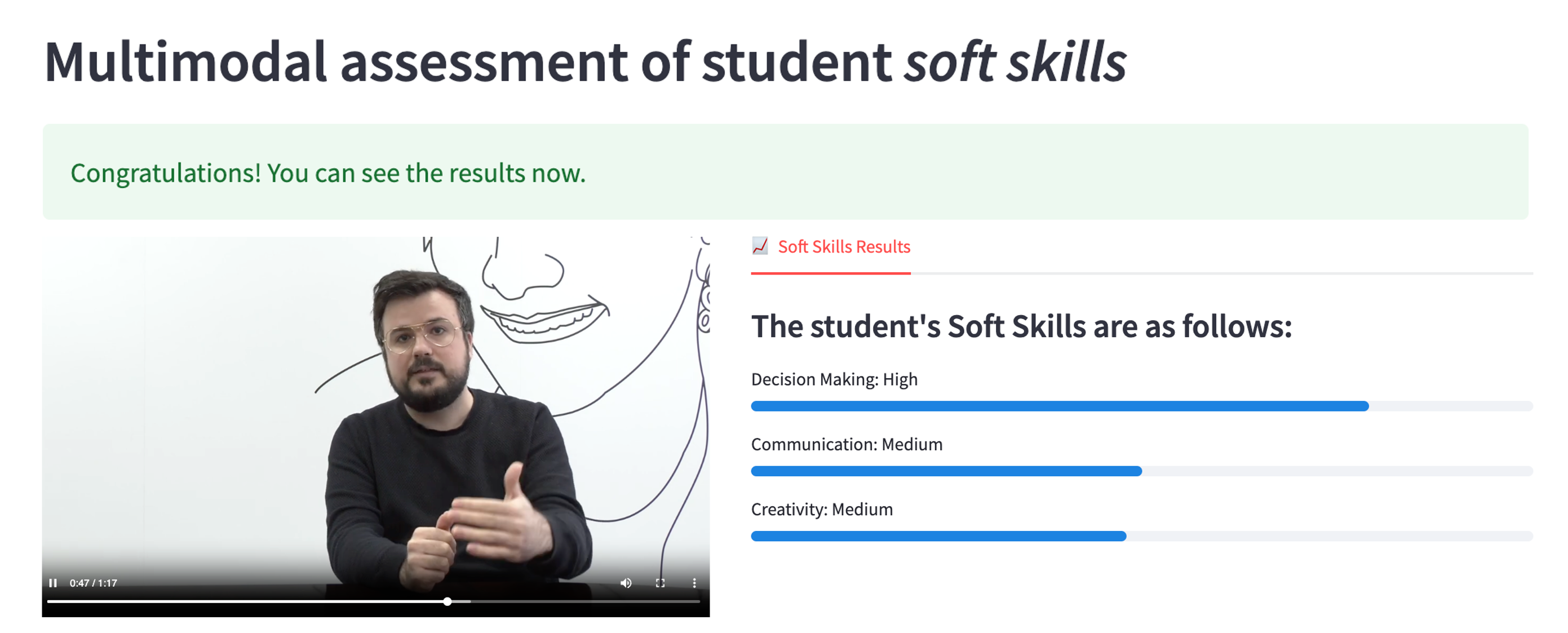}
    \caption{Tool Interface - Video of the learner's task and soft skills results}
    \label{fig:interface_1}
\end{figure}

The tool can be used through a web application where the teacher uploads the video and later visualises the results. The results interface comprises three elements: a video of the learner's task, results about soft skills, and a detailed report. Fig.~\ref{fig:interface_1} depicts the initial two elements. The video, situated on the left-hand side of the interface, is played automatically. The results of the soft skills assessment, displayed on the right, are represented by language labels and horizontal bars indicating the level of performance. These results are presented in a graphical and summarised form.

\begin{figure}[ht!]
    \centering
    \includegraphics[scale=0.14]{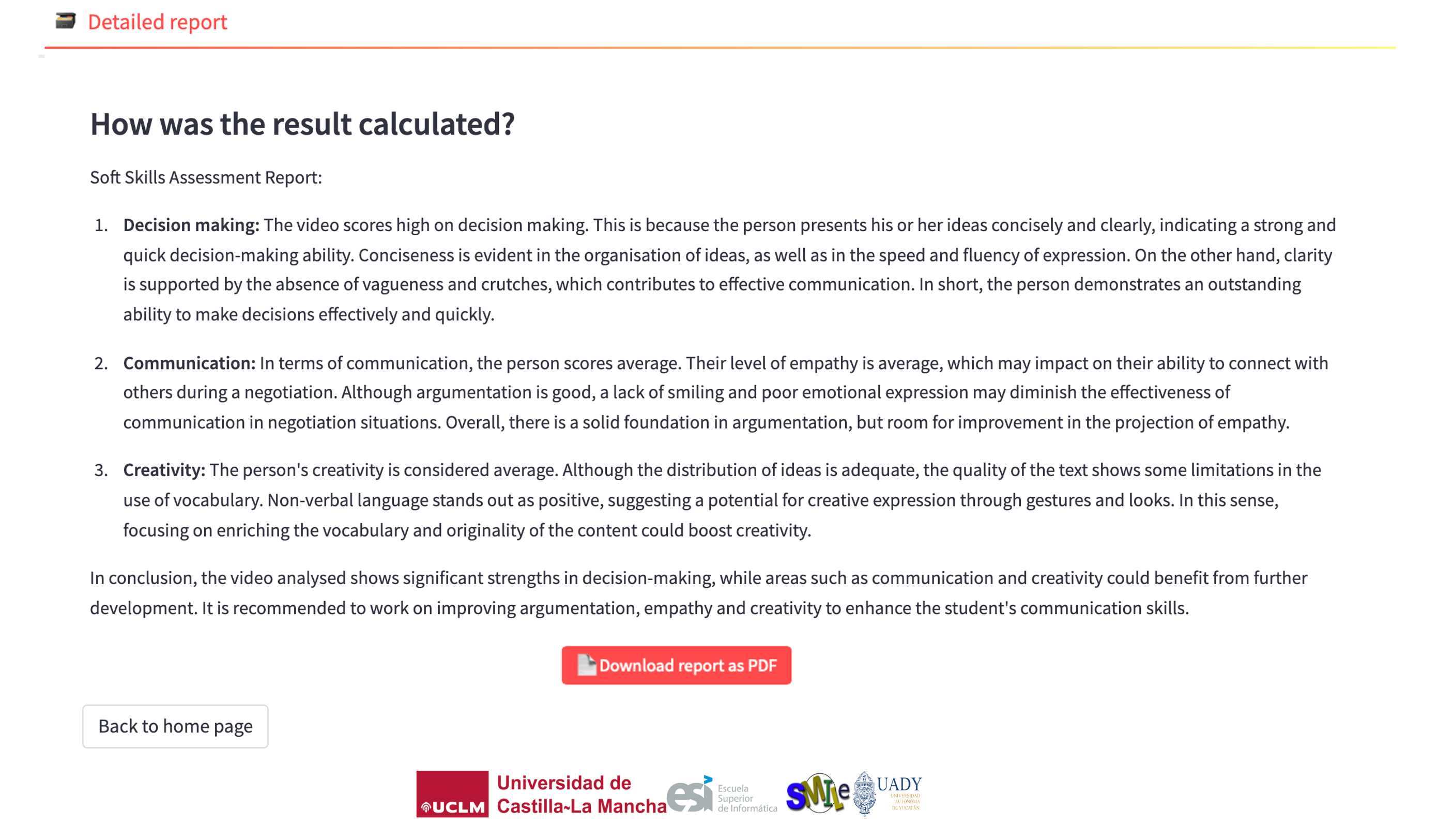}
    \caption{Tool Interface - Detailed report}
    \label{fig:interface_2}
\end{figure}

Fig. \ref{fig:interface_2} depicts the detailed report of the assessed video. This section provides a comprehensive evaluation breakdown, explaining the rationale behind the results presented in Fig. \ref{fig:interface_1}. The report includes detailed textual explanations for each assessed skill, providing insights into the learner’s performance and areas for improvement. These explanations aim to make the process used to generate the results fully transparent, allowing the user to understand how each evaluation was reached. 

Given the use of an AI engine, it is essential to guarantee that users have absolute confidence in the objectivity and accuracy of the results. This level of transparency helps to mitigate any uncertainties regarding the methods employed and reinforces trust in the fairness of the analysis. Furthermore, the comprehensive report is available for download in PDF format, allowing for further review at any time.

This combination of GLMP and LLM offers a dual-layered explainability mechanism that distinguishes this approach from conventional assessment frameworks. GLMP captures and organises complex behavioural data into structured, linguistically meaningful units, while the LLM transforms these structured outputs into fluent, pedagogically aligned narratives. This synergy ensures that not only are the system’s inferences grounded in interpretable fuzzy logic rules, but they are also communicated in a way that aligns with human reasoning and educational language. As a result, both technical and non-technical stakeholders can understand the rationale behind each assessment, reinforcing the transparency, trust, and pedagogical utility of the tool.

\section{Results}
\label{sec:results}
The following presents the analysis of the results obtained from the execution of the experiment, first for the group of Machine Learning students and subsequently for the Human-Computer Interaction students. This analysis also includes the instructor's assessment for each group.

\subsection{Machine Learning soft skill results}

The results obtained in the Machine Learning Subject in the 2022 edition (see Table~\ref{tab_soft-skills-group-A}) suggest that the type of evaluation (video vs. text) might significantly affect the performance in soft skills, particularly in decision-making. Students tend to perform better on text-based tasks than video-based tasks, which may be due to their familiarity with text-based tasks as opposed to the less frequent use of video.

On the other hand, evaluating communication skills does not offer excellent results using text-based tasks. The students' discourse in the essays is pretty standard, and then the conclusion of the evaluation methodology is always a medium score. Similar effects happened in creativity; the use of new vocabulary or end structures is not frequent in text, and this kind of expression in a video provides better evidence of the skill. Creativity seems slightly more expressed in the video task for some students (e.g., Student A5 shows Medium/High score in Task 1 and drops in subsequent tasks), suggesting that the dynamic format of video might spur creative expression better than static text for specific individuals. 

The discrepancy in performance observed across different media underscores the necessity for a diversified approach to evaluating soft skills, which can accommodate students' diverse strengths and expression styles. A mixed-method approach integrating video and text can offer a more comprehensive representation of a student's abilities.

\begin{table}[!htp]
    \centering
    \caption{Machine learning course 2022 edition soft skills results}
    \label{tab_soft-skills-group-A}
    \begin{tabular}{|c|*{9}{c|}}
     \hline
     \multirow{2}{*}{\textbf{Student}} & \multicolumn{3}{c}{\textbf{Decision-making}}  & \multicolumn{3}{c}{\textbf{Communication}}  & \multicolumn{3}{c|}{\textbf{Creativity}} \\
         & \textbf{T1 \includegraphics[height=8pt]{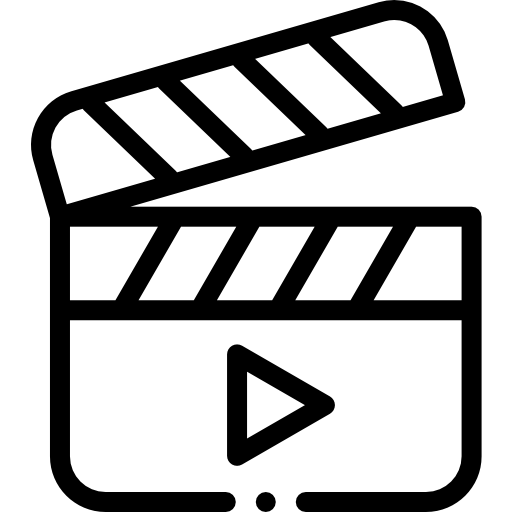}} & \textbf{T2 \includegraphics[height=8pt]{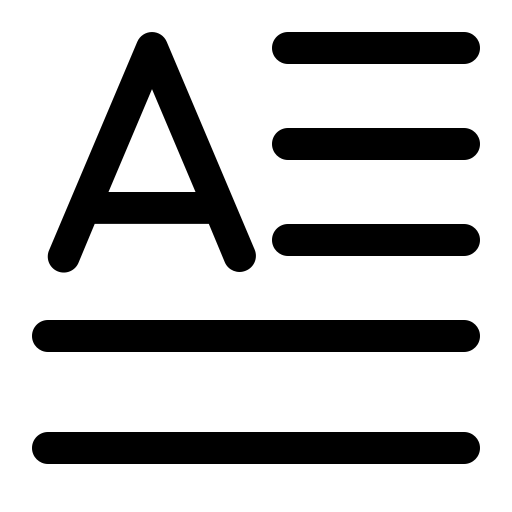}} & \textbf{T3 \includegraphics[height=8pt]{text_icon.png}} & \textbf{T1 \includegraphics[height=8pt]{video_icon.png}} & \textbf{T2 \includegraphics[height=8pt]{text_icon.png}} & \textbf{T3 \includegraphics[height=8pt]{text_icon.png}} & \textbf{T1 \includegraphics[height=8pt]{video_icon.png}} & \textbf{T2 \includegraphics[height=8pt]{text_icon.png}} & \textbf{T3 \includegraphics[height=8pt]{text_icon.png}} \\
        \hline 
        A1  & \cellcolor{Medium}M & \cellcolor{High}H & \cellcolor{LowMedium}L/M & \cellcolor{Low}L & \cellcolor{Medium}M & \cellcolor{Medium}M & \cellcolor{Medium}M & \cellcolor{Medium}M & \cellcolor{Medium}M  \\
        A2  & \cellcolor{LowMedium}L/M & \cellcolor{High}H & \cellcolor{MediumHigh}M/H & \cellcolor{Medium}M & \cellcolor{Medium}M & \cellcolor{Medium}M & \cellcolor{Low}L & \cellcolor{Medium}M & \cellcolor{LowMedium}L/M  \\
        A3  & \cellcolor{LowMedium}L/M & \cellcolor{High}H & \cellcolor{High}H & \cellcolor{Medium}M & \cellcolor{Medium}M & \cellcolor{Medium}M & \cellcolor{Medium}M & \cellcolor{Medium}M & \cellcolor{MediumHigh}M/H \\
        A4  & \cellcolor{Low}L & \cellcolor{High}H & \cellcolor{High}H & \cellcolor{Low}L & \cellcolor{Medium}M & \cellcolor{Medium}M & \cellcolor{LowMedium}L/M & \cellcolor{LowMedium}L/M & \cellcolor{Low}L \\
        A5  & \cellcolor{Medium}M & \cellcolor{High}H & \cellcolor{MediumHigh}M/H & \cellcolor{Medium}M & \cellcolor{Medium}M & \cellcolor{Medium}M & \cellcolor{MediumHigh}M/H & \cellcolor{Medium}M & \cellcolor{LowMedium}L/M \\
        A6  & \cellcolor{Low}L & \cellcolor{LowMedium}L/M & \cellcolor{High}H & \cellcolor{Low}L & \cellcolor{Medium}M & \cellcolor{Medium}M & \cellcolor{Low}L & \cellcolor{Medium}M & \cellcolor{Medium}M \\
        A7  & \cellcolor{LowMedium}L/M & \cellcolor{LowMedium}L/M & \cellcolor{LowMedium}L/M & \cellcolor{Medium}M & \cellcolor{Medium}M & \cellcolor{Medium}M & \cellcolor{Medium}M & \cellcolor{LowMedium}L/M & \cellcolor{Low}L \\
        A8  & \cellcolor{Medium}M & \cellcolor{MediumHigh}M/H & \cellcolor{Medium}M & \cellcolor{Medium}M & \cellcolor{Medium}M & \cellcolor{Medium}M & \cellcolor{LowMedium}L/M & \cellcolor{Low}L & \cellcolor{Low}L \\
        A9  & \cellcolor{LowMedium}L/M & \cellcolor{High}H & \cellcolor{High}H & \cellcolor{Low}L & \cellcolor{Medium}M & \cellcolor{Medium}M & \cellcolor{Medium}M & \cellcolor{Medium}M & \cellcolor{Low}L \\
        A10 & \cellcolor{High}H & \cellcolor{High}H & \cellcolor{High}H & \cellcolor{Medium}M & \cellcolor{Medium}M & \cellcolor{Medium}M & \cellcolor{MediumHigh}M/H & \cellcolor{Medium}M & \cellcolor{LowMedium}L/M \\
        \hline 
        \multicolumn{10}{|c|}{\textbf{L:} Low \textbf{L/M:} Low/Medium \textbf{M:} Medium \textbf{M/H:} Medium/High \textbf{H:} High} \\
        \hline
    \end{tabular}
\end{table}

In the 2023 edition of the Machine Learning (see Table~\ref{tab_soft-skills-group-B}), from the point of view of the tasks, Task 1 saw a range of performances with a tendency towards medium scores, and a few students achieving higher in creativity, perhaps reflecting initial enthusiasm or a novel approach to the video format. In Task 2, there was a descent in performance across most skills, particularly in decision-making and creativity, suggesting an increase in task difficulty. By Task 3, the communication scores for some students are high, which could indicate a gradual adaptation to expressing themselves via video. However, decision-making and creativity scores remained unchanged, reflecting ongoing challenges in these areas. 

\begin{table}[!htp]
    \centering
    \caption{Machine learning course 2023 edition soft skills results}
    \label{tab_soft-skills-group-B}
    \begin{tabular}{|c|*{9}{c|}}
     \hline
     \multirow{2}{*}{\textbf{Student}} & \multicolumn{3}{c}{\textbf{Decision-making}}  & \multicolumn{3}{c}{\textbf{Communication}}  & \multicolumn{3}{c|}{\textbf{Creativity}} \\
         & \textbf{T1 \includegraphics[height=8pt]{video_icon.png}} & \textbf{T2\includegraphics[height=8pt]{video_icon.png}} & \textbf{T3 \includegraphics[height=8pt]{video_icon.png}} & \textbf{T1 \includegraphics[height=8pt]{video_icon.png}} & \textbf{T2 \includegraphics[height=8pt]{video_icon.png}} & \textbf{T3 \includegraphics[height=8pt]{video_icon.png}} & \textbf{T1 \includegraphics[height=8pt]{video_icon.png}} & \textbf{T2 \includegraphics[height=8pt]{video_icon.png}} & \textbf{T3 \includegraphics[height=8pt]{video_icon.png}} \\
        \hline 
        B1  & \cellcolor{Medium}M      & \cellcolor{Low}L        & \cellcolor{Medium}M     & \cellcolor{Medium}M      & \cellcolor{Low}L        & \cellcolor{Medium}M      & \cellcolor{High}H        & \cellcolor{Low}L        & \cellcolor{Medium}M      \\
        B2  & \cellcolor{Medium}M      & \cellcolor{Medium}M     & \cellcolor{Medium}M     & \cellcolor{Low}L         & \cellcolor{Low}L        & \cellcolor{High}H        & \cellcolor{Medium}M      & \cellcolor{Medium}M     & \cellcolor{LowMedium}L/M  \\
        B3  & \cellcolor{Low}L         & \cellcolor{Medium}M     & \cellcolor{Medium}M     & \cellcolor{LowMedium}L/M  & \cellcolor{Low}L        & \cellcolor{MediumHigh}M/H & \cellcolor{Low}L         & \cellcolor{Low}L        & \cellcolor{Medium}M      \\
        B4  & \cellcolor{Low}L         & \cellcolor{Low}L        & \cellcolor{Low}L        & \cellcolor{Low}L         & \cellcolor{Low}L        & \cellcolor{LowMedium}L/M  & \cellcolor{Low}L         & \cellcolor{Low}L        & \cellcolor{LowMedium}L/M  \\
        B5  & \cellcolor{MediumHigh}M/H & \cellcolor{LowMedium}L/M & \cellcolor{Medium}M     & \cellcolor{High}H        & \cellcolor{Medium}M     & \cellcolor{High}H        & \cellcolor{MediumHigh}M/H & \cellcolor{Medium}M     & \cellcolor{MediumHigh}M/H \\
        B6  & \cellcolor{Low}L         & \cellcolor{Low}L        & \cellcolor{LowMedium}L/M & \cellcolor{MediumHigh}M/H & \cellcolor{Low}L        & \cellcolor{Low}L         & \cellcolor{High}H        & \cellcolor{High}H       & \cellcolor{LowMedium}L/M  \\
        B7  & \cellcolor{Medium}M      & \cellcolor{Low}L        & \cellcolor{Low}L        & \cellcolor{Medium}M      & \cellcolor{Low}L        & \cellcolor{Medium}M      & \cellcolor{Medium}M      & \cellcolor{Low}L        & \cellcolor{LowMedium}L/M  \\
        B8  & \cellcolor{Low}L         & \cellcolor{LowMedium}L/M & \cellcolor{Low}L        & \cellcolor{Low}L         & \cellcolor{Medium}M     & \cellcolor{Medium}M      & \cellcolor{Low}L         & \cellcolor{LowMedium}L/M & \cellcolor{Low}L         \\
        B9  & \cellcolor{LowMedium}L/M  & \cellcolor{Low}L        & \cellcolor{Low}L        & \cellcolor{Medium}M      & \cellcolor{Medium}M     & \cellcolor{Medium}M      & \cellcolor{High}H        & \cellcolor{Medium}M     & \cellcolor{Medium}M      \\
        B10 & \cellcolor{Low}L         & \cellcolor{Low}L        & \cellcolor{Low}L        & \cellcolor{Low}L         & \cellcolor{Low}L        & \cellcolor{Low}L         & \cellcolor{Low}L         & \cellcolor{LowMedium}L/M & \cellcolor{Low}L         \\
        B11 & \cellcolor{Low}L         & \cellcolor{Medium}M     & \cellcolor{Low}L        & \cellcolor{Low}L         & \cellcolor{Medium}M     & \cellcolor{Low}L         & \cellcolor{LowMedium}L/M  & \cellcolor{LowMedium}L/M & \cellcolor{Low}L         \\
        B12 & \cellcolor{LowMedium}L/M  & \cellcolor{Low}L        & \cellcolor{Low}L        & \cellcolor{Low}L         & \cellcolor{Low}L        & \cellcolor{Low}L         & \cellcolor{LowMedium}L/M  & \cellcolor{Low}L        & \cellcolor{Low}L         \\
        B13 & \cellcolor{LowMedium}L/M  & \cellcolor{Low}L        & \cellcolor{Low}L        & \cellcolor{Medium}M      & \cellcolor{High}H       & \cellcolor{High}H        & \cellcolor{Medium}M      & \cellcolor{Low}L        & \cellcolor{LowMedium}L/M  \\
        B14 & \cellcolor{LowMedium}L/M  & \cellcolor{Low}L        & \cellcolor{Low}L        & \cellcolor{Medium}M      & \cellcolor{Medium}M     & \cellcolor{High}H        & \cellcolor{Medium}M      & \cellcolor{Medium}M     & \cellcolor{LowMedium}L/M  \\
        B15 & \cellcolor{High}H        & \cellcolor{Low}L        & \cellcolor{High}H       & \cellcolor{Low}L         & \cellcolor{LowMedium}L/M & \cellcolor{Low}L         & \cellcolor{High}H        & \cellcolor{Low}L        & \cellcolor{Medium}M      \\
        B16 & \cellcolor{Medium}M      & \cellcolor{Low}L        & \cellcolor{Medium}M     & \cellcolor{High}H        & \cellcolor{Medium}M     & \cellcolor{Medium}M      & \cellcolor{LowMedium}L/M  & \cellcolor{Low}L        & \cellcolor{LowMedium}L/M  \\
        B17 & \cellcolor{Medium}M      & \cellcolor{Low}L        & \cellcolor{Low}L        & \cellcolor{Medium}M      & \cellcolor{Medium}M     & \cellcolor{LowMedium}L/M  & \cellcolor{Medium}M      & \cellcolor{Medium}M     & \cellcolor{Medium}M      \\
        B18 & \cellcolor{LowMedium}L/M  & \cellcolor{Low}L        & \cellcolor{LowMedium}L/M & \cellcolor{Low}L         & \cellcolor{Low}L        & \cellcolor{LowMedium}L/M  & \cellcolor{Medium}M      & \cellcolor{LowMedium}L/M & \cellcolor{LowMedium}L/M \\
        \hline
        \multicolumn{10}{|c|}{\textbf{L:} Low \textbf{L/M:} Low/Medium \textbf{M:} Medium \textbf{M/H:} Medium/High \textbf{H:} High} \\ \hline
    \end{tabular}
\end{table}

From the point of view of skills, decision-making scores ranged from low to medium across the tasks, with no significant improvements noted as the tasks progressed. Communication presents a different behaviour, with some students improving over time, particularly by the third task, suggesting a possible adaptation to the video format. Creativity scores were predominantly low to medium, with rare high scores indicating that students found it challenging to express creativity consistently in the video. These trends suggest that while some students may excel in specific skills, others might need additional support. Therefore, video-based evaluations for soft skills, as seen in the comparative analysis of the 2022 and 2023 Machine Learning course editions, offer significant advantages over text-based assessments, particularly in capturing non-verbal cues such as body language and tone, which are essential for a comprehensive understanding of communication skills. Video formats provide a more practical environment for assessing decision-making or creativity. Nevertheless, this evaluation method has shortcomings, such as potential initial discomfort with the medium. 

\subsection{Human-Computer Interaction soft skill results}
The assessment of soft skills through video tasks in the Human Interaction Computer in group A (see Table \ref{tab_soft-skills-group-CA}) has revealed notable discrepancies in student performance across soft skills. For example, student CA2 demonstrates proficiency in a range of tasks, consistently achieving between high and medium scores. In contrast, student CA3 tends to exhibit challenges in meeting the same level of performance, as evidenced by their lower scores. This performance diversity indicates that students demonstrate different levels of competence in these soft skills.

\begin{table}[!htp]
    \caption{Interaction Human-Computer course group A soft skills results}
    \label{tab_soft-skills-group-CA}
    \centering
    \begin{tabular}{|c|*{12}{c|}}
     \hline
     \multirow{2}{*}{\textbf{Student}} & \multicolumn{3}{c}{\textbf{Decision-making}}  & \multicolumn{3}{c}{\textbf{Communication}}  & \multicolumn{3}{c|}{\textbf{Creativity}} \\
         & \textbf{T1 \includegraphics[height=8pt]{video_icon.png}} & \textbf{T2\includegraphics[height=8pt]{video_icon.png}} & \textbf{T3 \includegraphics[height=8pt]{video_icon.png}} & \textbf{T1 \includegraphics[height=8pt]{video_icon.png}} & \textbf{T2 \includegraphics[height=8pt]{video_icon.png}} & \textbf{T3 \includegraphics[height=8pt]{video_icon.png}} & \textbf{T1 \includegraphics[height=8pt]{video_icon.png}} & \textbf{T2 \includegraphics[height=8pt]{video_icon.png}} & \textbf{T3 \includegraphics[height=8pt]{video_icon.png}} \\
        \hline 
        CA1  & \cellcolor{Medium}M     & \cellcolor{High}H        & \cellcolor{Low}L        & \cellcolor{Medium}M     & \cellcolor{LowMedium}L/M & \cellcolor{Low}L        & \cellcolor{Low}L         & \cellcolor{LowMedium}L/M & \cellcolor{Low}L        \\
        CA2  & \cellcolor{High}H       & \cellcolor{MediumHigh}M/H & \cellcolor{High}H       & \cellcolor{Medium}M     & \cellcolor{LowMedium}L/M & \cellcolor{High}H       & \cellcolor{Medium}M      & \cellcolor{LowMedium}L/M & \cellcolor{High}H       \\
        CA3  & \cellcolor{Low}L        & \cellcolor{Low}L         & \cellcolor{Low}L        & \cellcolor{Medium}M     & \cellcolor{Low}L        & \cellcolor{LowMedium}L/M & \cellcolor{Medium}M      & \cellcolor{Low}L        & \cellcolor{Medium}M     \\
        CA4  & \cellcolor{High}H       & \cellcolor{Medium}M      & \cellcolor{Medium}M     & \cellcolor{Low}L        & \cellcolor{Medium}M     & \cellcolor{Medium}M     & \cellcolor{MediumHigh}M/H & \cellcolor{Low}L        & \cellcolor{Medium}M     \\
        CA5  & \cellcolor{Medium}M     & \cellcolor{Low}L         & \cellcolor{Low}L        & \cellcolor{Low}L        & \cellcolor{LowMedium}L/M & \cellcolor{Low}L        & \cellcolor{Medium}M      & \cellcolor{LowMedium}L/M & \cellcolor{Low}L        \\
        CA6  & \cellcolor{Medium}M     & \cellcolor{High}H        & \cellcolor{Low}L        & \cellcolor{High}H       & \cellcolor{Medium}M     & \cellcolor{Low}L        & \cellcolor{MediumHigh}M/H & \cellcolor{High}H       & \cellcolor{Medium}M     \\
        CA7  & \cellcolor{High}H       & \cellcolor{Medium}M      & \cellcolor{LowMedium}L/M & \cellcolor{Medium}M     & \cellcolor{High}H       & \cellcolor{LowMedium}L/M & \cellcolor{MediumHigh}M/H & \cellcolor{LowMedium}L/M & \cellcolor{LowMedium}L/M \\
        CA8  & \cellcolor{High}H       & \cellcolor{MediumHigh}M/H & \cellcolor{High}H       & \cellcolor{High}H       & \cellcolor{LowMedium}L/M & \cellcolor{Medium}M     & \cellcolor{Medium}M      & \cellcolor{Low}L        & \cellcolor{Medium}M     \\
        CA9  & \cellcolor{High}H       & \cellcolor{Low}L         & \cellcolor{Low}L        & \cellcolor{LowMedium}L/M & \cellcolor{Low}L        & \cellcolor{Low}L        & \cellcolor{Medium}M      & \cellcolor{Low}L        & \cellcolor{Low}L        \\
        CA10 & \cellcolor{LowMedium}L/M & \cellcolor{Medium}M      & \cellcolor{High}H       & \cellcolor{Low}L        & \cellcolor{Medium}M     & \cellcolor{Low}L        & \cellcolor{Low}L         & \cellcolor{LowMedium}L/M & \cellcolor{Medium}M \\
        \hline
        \multicolumn{10}{|c|}{\textbf{L:} Low \textbf{L/M:} Low/Medium \textbf{M:} Medium \textbf{M/H:} Medium/High \textbf{H:} High} \\
        \hline
    \end{tabular}
\end{table}

The analysis of the obtained results reveals a considerable range of abilities among students in decision-making. For example, students CA2 and CA4 demonstrate proficiency in these skills, whereas students CA3 and CA5 exhibit considerable deficiencies. Communication scores are more evenly distributed, with some students, such as CA8, demonstrating high proficiency, while others, like CA9, exhibit less developed skills. In contrast, creativity scores are generally lower across the board, indicating that this skill presents particular challenges for many students. Furthermore, the absence of a correlation between different soft skills indicates that strengths in one area do not necessarily translate to others, emphasising the necessity to evaluate each skill independently.

In the experiment for group B (Table \ref{tab_soft-skills-group-CB}), in which Task 2 was conducted via text and the remaining tasks via video, the results demonstrate a range of performance across the soft skills. Decision-making abilities were generally strong, particularly in Tasks 1 and 2, where students CB9, CB10, and CB11 demonstrated consistently high performance, indicating proficiency in decision-making across video and text formats. However, some students, such as CB1 and CB3, exhibit lower or medium scores across all categories, indicating less consistent decision-making across different task types. The communication skills of some students may not vary significantly between text and video formats. Nevertheless, a few students, such as CB2 and CB6, consistently demonstrated medium scores across all tasks, indicating strong communication skills regardless of the format. Overall, creativity scores tend to be lower, particularly in Task 3.

\begin{table}[!htp]
    \caption{Interaction Human-Computer course group B soft skills results}
    \label{tab_soft-skills-group-CB}
    \centering
    \begin{tabular}{|c|*{9}{c|}}
     \hline
     \multirow{2}{*}{\textbf{Student}} & \multicolumn{3}{c}{\textbf{Decision-making}}  & \multicolumn{3}{c}{\textbf{Communication}}  & \multicolumn{3}{c|}{\textbf{Creativity}} \\
         & \textbf{T1 \includegraphics[height=8pt]{video_icon.png}} & \textbf{T2 \includegraphics[height=8pt]{text_icon.png}} & \textbf{T3 \includegraphics[height=8pt]{video_icon.png}} & \textbf{T1 \includegraphics[height=8pt]{video_icon.png}} & \textbf{T2 \includegraphics[height=8pt]{text_icon.png}} & \textbf{T3 \includegraphics[height=8pt]{video_icon.png}} & \textbf{T1 \includegraphics[height=8pt]{video_icon.png}} & \textbf{T2 \includegraphics[height=8pt]{text_icon.png}} & \textbf{T3 \includegraphics[height=8pt]{video_icon.png}} \\
        \hline 
        CB1  & \cellcolor{Medium}M      & \cellcolor{LowMedium}L/M  & \cellcolor{LowMedium}L/M  & \cellcolor{Medium}M     & \cellcolor{Medium}M & \cellcolor{Low}L        & \cellcolor{MediumHigh}M/H & \cellcolor{LowMedium}L/M & \cellcolor{LowMedium}L/M  \\
        CB2  & \cellcolor{MediumHigh}M/H & \cellcolor{MediumHigh}M/H & \cellcolor{High}H        & \cellcolor{Low}L        & \cellcolor{Medium}M & \cellcolor{Medium}M     & \cellcolor{Medium}M      & \cellcolor{Low}L        & \cellcolor{High}H        \\
        CB3  & \cellcolor{LowMedium}L/M  & \cellcolor{LowMedium}L/M  & \cellcolor{LowMedium}L/M  & \cellcolor{Medium}M     & \cellcolor{Medium}M & \cellcolor{Low}L        & \cellcolor{Medium}M      & \cellcolor{LowMedium}L/M & \cellcolor{Low}L         \\
        CB4  & \cellcolor{LowMedium}L/M  & \cellcolor{Medium}M      & \cellcolor{Low}L         & \cellcolor{High}H       & \cellcolor{Medium}M & \cellcolor{Low}L        & \cellcolor{LowMedium}L/M  & \cellcolor{LowMedium}L/M & \cellcolor{MediumHigh}M/H \\
        CB5  & \cellcolor{MediumHigh}M/H & \cellcolor{MediumHigh}M/H & \cellcolor{High}H        & \cellcolor{Medium}M     & \cellcolor{Medium}M & \cellcolor{Medium}M     & \cellcolor{Medium}M      & \cellcolor{LowMedium}L/M & \cellcolor{Medium}M      \\
        CB6  & \cellcolor{MediumHigh}M/H & \cellcolor{Medium}M      & \cellcolor{LowMedium}L/M  & \cellcolor{Medium}M     & \cellcolor{Medium}M & \cellcolor{Medium}M     & \cellcolor{High}H        & \cellcolor{LowMedium}L/M & \cellcolor{Low}L         \\
        CB7  & \cellcolor{MediumHigh}M/H & \cellcolor{Low}L         & \cellcolor{Medium}M      & \cellcolor{LowMedium}L/M & \cellcolor{Medium}M & \cellcolor{LowMedium}L/M & \cellcolor{Medium}M      & \cellcolor{LowMedium}L/M & \cellcolor{Low}L         \\
        CB8  & \cellcolor{MediumHigh}M/H & \cellcolor{MediumHigh}M/H & \cellcolor{Low}L         & \cellcolor{Medium}M     & \cellcolor{Medium}M & \cellcolor{Low}L        & \cellcolor{Medium}M      & \cellcolor{Medium}M     & \cellcolor{Low}L         \\
        CB9  & \cellcolor{High}H        & \cellcolor{High}H        & \cellcolor{MediumHigh}M/H & \cellcolor{Medium}M     & \cellcolor{Medium}M & \cellcolor{Medium}M     & \cellcolor{High}H        & \cellcolor{LowMedium}L/M & \cellcolor{MediumHigh}M/H \\
        CB10 & \cellcolor{High}H        & \cellcolor{Medium}M      & \cellcolor{LowMedium}L/M  & \cellcolor{High}H       & \cellcolor{Medium}M & \cellcolor{LowMedium}L/M & \cellcolor{MediumHigh}M/H & \cellcolor{LowMedium}L/M & \cellcolor{LowMedium}L/M  \\
        CB11 & \cellcolor{High}H        & \cellcolor{MediumHigh}M/H & \cellcolor{LowMedium}L/M  & \cellcolor{Medium}M     & \cellcolor{Medium}M & \cellcolor{Medium}M     & \cellcolor{Medium}M      & \cellcolor{LowMedium}L/M & \cellcolor{Low}L   \\
        \hline
        \multicolumn{10}{|c|}{\textbf{L:} Low \textbf{L/M:} Low/Medium \textbf{M:} Medium \textbf{M/H:} Medium/High \textbf{H:} High} \\
        \hline
    \end{tabular}
\end{table}

Both experiments demonstrate that the task format affects the evaluation of soft skills, with decision-making and communication being more adaptable across video and text tasks. Decision-making consistently demonstrates robust outcomes, whereas communication exhibits stability, i.e., less susceptible to format-related influences. However, the evaluation of creativity was a challenge in both formats, with students scoring lower overall. 

\subsection{Correlation Analysis Between Soft Skill Scores and Teacher Ratings}
To strengthen the empirical validation of the proposed framework, a correlation analysis was conducted between the soft skill scores automatically generated by the system and the evaluations assigned by course instructors. Only tasks fully based on video submissions were considered, as these allowed for comprehensive multimodal analysis. In the case of Task 2 from the Human-Computer Interaction group, only the 10 students from Group A—who completed all tasks in video format—were included in the analysis. The individual ratings per task are shown in Table \ref{tab_grades}.

\begin{table}[!htp]
    \caption{Correlation of the skills with the teacher ratings}
    \label{tab_correlation}
    \centering
    
    \begin{tabular}{|l|c|c|c|c|}
     \hline
     \textbf{Group} & \textbf{Task} & \textbf{Decision-making} & \textbf{Communication} & \textbf{Creativity} \\ \hline
     Machine learning 2022 & 1 & 0.74 & 0.57 & 0.94 \\ \hline
     \multirow{3}{*}{Machine learning 2023} & 1 & 0.78 & 0.57 & 0.93 \\
     & 2 & 0.17 & 0.86 & 0.6 \\
     & 3 & 0.87 & 0.51 & 0.73 \\ \hline
     \multirow{3}{*}{Interaction Human-Computer} & 1 & 0.89 & 0.36 & 0.69 \\
     & 2 & 0.63 & 0.78 & 0.91 \\
     & 3 & 0.82 & 0.84 & 0.75  \\ \hline
    \end{tabular}
\end{table}

Table~\ref{tab_correlation} presents the Pearson correlation coefficients for each soft skill across different tasks and groups. The results show generally strong positive correlations between the system-generated scores and the instructors' ratings, particularly in creativity and decision-making. For instance, Task 1 in the Machine Learning 2022 group yielded a very strong correlation for creativity (0.94), with decision-making also showing solid alignment (0.74). Similar trends were observed in the 2023 edition, where Task 1 and Task 3 reported high correlations in creativity (0.93) and decision-making (0.78), respectively. Communication correlations in these tasks were more moderate, typically ranging from 0.51 to 0.57.

These findings align with the pedagogical structure of the Machine Learning course, in which each task emphasised a different skill: creativity was particularly relevant in Task 1 (personal video presentation), communication was central to Task 2 (summary of a talk), and decision-making was targeted in Task 3 (report on key points and future directions). This alignment between task focus and skill correlation reinforces the construct validity of the proposed framework.

The Human-Computer Interaction group demonstrated similar alignment patterns. In this course, the instructor’s assessments were also guided by the specific learning objectives of each task. Decision-making was the primary focus in Task 1, which resulted in a high correlation of 0.89. Creativity was the central aspect in Task 2, where the system also showed strong alignment (0.91). Finally, Task 3 prioritised communication, and again the system’s output closely matched the instructor's evaluations (0.84). These results confirm the framework's capability to capture specific soft skills when the instructional focus is well defined.

Some variation in correlation values was observed across tasks, which can be attributed to the differing emphasis of each assignment. For example, the lower correlation in decision-making observed in Task 2 of the Machine Learning 2023 group (0.17) is consistent with the nature of that activity, which primarily focused on communication rather than decision-making. This result reinforces the importance of aligning task design with the targeted skill when interpreting automated assessments.

Overall, these findings provide empirical support for the validity of the proposed framework, demonstrating that the automated assessments are largely consistent with expert human judgement, particularly when task design is pedagogically aligned with the soft skill being evaluated. This analysis helps to address prior concerns regarding the reliability, accuracy, and generalisability of the system’s core claims.

\subsection{Satisfaction Evaluation}

Evaluating teacher satisfaction is essential to understanding the impact and effectiveness of the video-based soft skills assessment method. For this purpose, for the Machine Learning courses of 2022 (ML 2022) and 2023 (ML 2023) and Interaction Human-Computer (HU), which includes group A (HU-A) and group B (HU-B), a validated, standardized instrument was used to measure satisfaction rather than relying on ad-hoc questionnaires~\cite{roman2023evaluating} as can be seen in Table~\ref{tab_satisfaction_evaluation} and the details of which can be seen in Table \ref{tab_satisfaction_evaluation_explained}.

The analysis of teacher satisfaction with soft skill assessment in ML 2022 and ML 2023 reveals that teachers were consistently satisfied with the assessments' clarity, consistency, and fairness across both courses. Both groups also appreciated the support provided to learners in processing feedback. This consistency suggests that the evaluation framework is technically sound and treats all participants equally, ensuring a positive perception of the assessment process.
However, there are some significant differences between both groups. The ML 2023 course showed progress in using assessments to promote student accountability and reflection and to gather valuable data to adjust teaching, improvements that were not evident in ML 2022. Despite these improvements, both years need more opportunities for learners to close the gap between current and desired performance. This suggests an important area for future development in the course structure.

\begin{table}[!htp]
    \caption{Teacher Satisfaction Evaluation}
    \label{tab_satisfaction_evaluation}
    \centering
    \small
    \begin{tabular}{|p{6cm}|c|c|c|}
     \hline
     \textbf{Measure} & \textbf{ML 2022} & \textbf{ML 2023} & \textbf{HU} \\
     \hline
     Transparency and Consistency & \ding{51} & \ding{51} & \ding{51} \\ \hline

     Impartial Assessment & \ding{51} & \ding{51} & \ding{51} \\ \hline

     Feedback Support & \ding{51} & \ding{51} & \ding{51} \\ \hline

     Accountability and Reflection &  & \ding{51} & \ding{51} \\ \hline

     Performance Alignment &  &  & \ding{51} \\ \hline

     Motivation and Feedback Value & \ding{51} & \ding{51} & \ding{51} \\ \hline

     Instructional Insights &  & \ding{51} & \ding{51} \\ \hline
    \end{tabular}
\end{table}

Analysis of student feedback (see Table~\ref{tab_satisfaction_feedback} and the related questions for the aspects to evaluate in Table \ref{tab_satisfaction_feedback_explained}) from different course setups, using different combinations of video and text tasks, reveals distinct patterns in student satisfaction and the perceived effectiveness of these assessment tools. ML 2022 and HU-B, which use a mixture of video and text tasks, were positively rated for helping students understand the subject and prepare for theory exams. This result suggests that a hybrid approach combining video and text formats improves comprehension and exam readiness, a benefit not observed in the all-video task setups of ML 2023 and HU-A, where these attributes were not rated positively. 

On the other hand, courses that relied exclusively on video tasks, particularly ML 2023 and HU-A, received positive feedback regarding their ability to increase student interest and motivation. These groups found video tasks more engaging and motivating for students than mixed formats. The consistent satisfaction across all groups with logistical aspects suggests that course schedules and time allocation are well planned. In addition, the positive feedback about the rewarding nature of task completion in ML 2023 and HU-A underlines the engaging and satisfying experience that video tasks can provide. These findings demonstrate the potential of video tasks to increase student engagement. In contrast, a balanced combination with text tasks may provide a more balanced pedagogical benefit, supporting comprehension and interest. In conclusion, educators can consider combining video and text tasks to improve student engagement and learning outcomes.

\begin{table}[!htp]
    \caption{Students Satisfaction and Feedback}
    \label{tab_satisfaction_feedback}
    \centering
    \small
    \begin{tabular}{|p{4cm}|c|c|c|c|}
     \hline
     \textbf{Aspect to Evaluate} & \textbf{ML 2022} & \textbf{ML 2023} & \textbf{HU-A} & \textbf{HU-B} \\ \hline

     Understanding of Subject & \ding{51} (90\%) &  &  & \ding{51} (91\%) \\ \hline
     
     Exam Preparation & \ding{51} (90\%) &  &  & \ding{51} (82\%) \\ \hline
     
     Interest in Subject & \ding{51} (80\%) & \ding{51} (94\%) & \ding{51} (90\%) & \\ \hline
     
     Participation Motivation &  & \ding{51} (100\%) & \ding{51} (80\%) & \\ \hline
     
     Didactic Resource Quality &  & \ding{51} (89\%) & \ding{51} (90\%) & \\ \hline
     
     Task Timing Satisfaction & \ding{51} (90\%) & \ding{51} (89\%) & \ding{51} (100\%) & \ding{51} (100\%) \\ \hline
     
     Task Completion Time & \ding{51} (100\%) & \ding{51} (94\%) & \ding{51} (90\%) & \ding{51} (91\%) \\ \hline
     
     Satisfaction from Rewards &  & \ding{51} (100\%) & \ding{51} (90\%) & \\ 
     \hline
    \end{tabular}
\end{table}

\section{Discussion and Limitations}
\label{sec:discussion_and_limitations}

The results of this study demonstrate the potential of multimodal analysis in assessing soft skills, particularly decision-making, creativity and communication among undergraduate students. Concerning RQ1, the study shows that incorporating a mix of video, audio and text data in the evaluation process enhances the ability to capture nuanced aspects of soft skills. In particular, video allows for assessing non-verbal cues such as facial expressions providing a more holistic assessment than traditional text-based methods. However, the effectiveness of these assessments depends on the medium used, as video tasks often reveal strengths and challenges that are not apparent in text-based assessments. For example, students tend to perform better in decision-making when given written tasks, while creativity is more visible in dynamic formats such as video.

About RQ2, integrating linguistic models and artificial intelligence is relevant in interpreting verbal and non-verbal cues. The AI-driven multimodal tool demonstrated a robust ability to analyse speech patterns, body language and emotional expressions, thereby improving the accuracy of the soft skills assessment. The application of natural language processing techniques enabled the tool to assess the content of students' verbal expressions, while deep learning models accurately detected facial expressions and eye focus. It should be noted, however, that students' moods and emotional states can impact the output of such assessments, introducing an element of variability into the assessment results. 

For RQ3, the integration of deep learning and fuzzy logic led to a notable enhancement in the precision and comprehensibility of the assessments. The application of fuzzy logic enabled the resolution of uncertainties inherent to the evaluation of emotional and behavioural cues. This resulted in the generation of clear, human-readable linguistic labels, thereby enhancing transparency for stakeholders. Nevertheless, there are still some limitations to the system, including its reliance on video-based tasks, which may not fully capture the full range of student abilities, and issues such as students' familiarity with video tasks. A hybrid assessment strategy, which incorporates both video and text-based tasks, may prove a more comprehensive means of evaluating students' skills.

Although the results are promising, there are several limitations to this study. Firstly, students' emotional state during the video tasks can significantly influence the results, leading to variability that does not reflect their abilities. This variability indicates the need for preparatory sessions or resources to help students adapt to video formats and demonstrate their skills more effectively. Furthermore, while valuable, the study's reliance on video-based assessment may not be appropriate for all learning environments or student preferences. Technical limitations such as video quality, internet connectivity, and students' comfort in front of the camera may also affect the results. Finally, while deep learning and fuzzy logic enhanced the assessment process, the models may require further refinement to address cultural and linguistic differences, particularly in international contexts. For example, students at the Autonomous University of Yucatan utilise a distinctive form of Spanish influenced by the Maya language, which affects lexicon, phonology, and syntax.

From an ethical standpoint, the use of AI for soft skills assessment demands careful attention to privacy, fairness, and transparency. The system ensures anonymity and secure handling of sensitive multimodal data, while the GLMP-LLM integration enhances explainability, allowing stakeholders to understand how scores are derived. However, potential biases in pretrained models and cultural variability in behavioural expressions require ongoing evaluation to ensure equitable and responsible use of the tool.

\section{Conclusions}
\label{sec:conclusions}

This study presents an advanced framework for assessing soft skills in educational contexts by combining linguistic models of perception with multimodal analysis. The integration of video, audio and text data enables a more comprehensive assessment of decision-making, creativity and communication, capturing both verbal and non-verbal cues such as facial expressions, tone of voice and emotional states. This approach provides a solution to the shortcomings of conventional text-based assessments, which are often inadequate for capturing the intricacies of human interactions. The use of fuzzy logic further strengthens the framework by providing nuanced interpretations of soft skills, improving the accuracy and interpretability of the assessment.

The results suggest that the combination of different artificial intelligence methods provides a scalable and dynamic solution for personalised feedback in educational settings. However, video-based assessments can introduce performance variability, influenced by students' comfort levels, emotional states and technical constraints, which may affect the reliability of results. Additionally, fuzzy logic enhances the interpretability of human behaviours, offering a powerful tool to capture the subtleties of complex interpersonal skills. While some subjectivity is inevitable in assessing such nuanced attributes, fuzzy logic provides a structured approach that brings clarity and depth to understanding these skills, making evaluations more insightful and adaptable.

Future work will focus on extending the adaptability and applicability of the framework. Hybrid assessment methods, combining video with traditional text assessment, are proposed to accommodate different learning styles and provide a more rounded view of students' skills. The implementation of structured feedback mechanisms could provide students with real-time insights, promoting engagement and growth. In addition, adapting the model for cultural and linguistic diversity using diverse data sets would increase its effectiveness in international educational settings. Further investigation into the incorporation of physiological signals could enhance the system's ability to assess emotional states, providing a richer understanding of students' soft skill development in different educational contexts.

\section*{Statements \& Declarations}

\subsection*{Funding} The Spanish Government has partially supported this work under the grant SAFER: PID2019-104735RB-C42 (ERA/ERDF, EU), and project PLEC2021-007681 funded by MCIN/AEI /10.13039/501100011033 and by the European Union NextGeneration EU/ PRTR.

\subsection*{Competing interests} The authors declare that they have no conflict of interest. The authors have no relevant financial or non-financial interests to disclose.

\subsection*{Authors' contributions} All authors contributed to the study's conception and design. Jared D.T. Guerrero-Sosa, Francisco P. Romero and Víctor Hugo Menéndez-Domínguez performed conceptualization, data collection and formal analysis. Jesus Serrano-Guerrero, Andres Montoro-Montarroso and Jose A. Olivas performed the experiment design and state-of-the-art review. 
Jared D.T. Guerrero-Sosa wrote the manuscript's first draft, and all authors commented on previous versions. All authors read and approved the final manuscript.

\bibliographystyle{unsrt}  
\bibliography{references}  

\begin{thebibliography}{10}

\bibitem{putra2020}
Arman~Syah Putra, Dewiana Novitasari, Masduki Asbari, Agus Purwanto, Joni Iskandar, Dhaniel Hutagalung, Suroso, and Yoyok Cahyono.
\newblock Examine relationship of soft skills, hard skills, innovation and performance: the mediation effect of organizational learning.
\newblock {\em International Journal of Science and Management Studies}, 3:27--43, 2020.

\bibitem{Ulfa2024}
Zahra~Maria Ulfa and Undang Rosidin.
\newblock Development of instrument soft skill assessment in problem-based learning to improve students’ collaboration skills, communication skills, and social skills.
\newblock {\em Jurnal Penelitian Pendidikan IPA}, 10(7):3681--3688, 2024.

\bibitem{Rasipuram2020}
Sowmya Rasipuram and Dinesh~Babu Jayagopi.
\newblock Automatic multimodal assessment of soft skills in social interactions: a review.
\newblock {\em Multimedia Tools and Applications}, 79:13037--13060, 2020.

\bibitem{mestre2023}
María~Angeles Mestre-Segarra.
\newblock A multimodal rhetorical analysis of video resumes.
\newblock {\em ESP Today}, 11(2):349--370, 2023.

\bibitem{Barthakur2023}
Abhinava Barthakur, Vitomir Kovanovic, Srecko Joksimovic, and Abelardo Pardo.
\newblock Challenges in assessments of soft skills: Towards unobtrusive approaches to measuring student success.
\newblock In {\em Unobtrusive Observations of Learning in Digital Environments: Examining Behavior, Cognition, Emotion, Metacognition and Social Processes Using Learning Analytics}, chapter~4, pages 53--71. Springer International Publishing, Cham, 2023.

\bibitem{Cheng2024}
Biqian Cheng, Yuping Liu, and Yunjian Jia.
\newblock Evaluation of students' performance during the academic period using the xg-boost classifier-enhanced aeo hybrid model.
\newblock {\em Expert Systems with Applications}, 238:122136, 2024.

\bibitem{Ngo2024}
Duong Ngo, Andy Nguyen, Belle Dang, and Ha~Ngo.
\newblock Facial expression recognition for examining emotional regulation in synchronous online collaborative learning.
\newblock {\em International Journal of Artificial Intelligence in Education}, 34:650--669, 2024.

\bibitem{deMorais2024}
Felipe De~Morais and Patricia~A Jaques.
\newblock The dynamics of brazilian students’ emotions in digital learning systems.
\newblock {\em International Journal of Artificial Intelligence in Education}, 34:519--544, 2024.

\bibitem{Guerrero-Sosa2023}
Jared D~T Guerrero-Sosa, Francisco~P Romero, Victor~H Menendez, Jesus Serrano-Guerrero, Jose~A Olivas, and Andres Montoro-Montarroso.
\newblock Granular linguistic model based multimodal data integration for automated evaluation of core soft skills.
\newblock In José Bravo and Gabriel Urzáiz, editors, {\em Proceedings of the 15th International Conference on Ubiquitous Computing \& Ambient Intelligence (UCAmI 2023)}, pages 292--303, Cham, 2023. Springer Nature Switzerland.

\bibitem{Feraco2023}
Tommaso Feraco, Dario Resnati, Davide Fregonese, Andrea Spoto, and Chiara Meneghetti.
\newblock An integrated model of school students’ academic achievement and life satisfaction. linking soft skills, extracurricular activities, self-regulated learning, motivation, and emotions.
\newblock {\em European Journal of Psychology of Education}, 38:109--130, 2023.

\bibitem{BORGES2024107395}
Giovana~Giardini Borges and Rogéria~Cristiane Gratão~de Souza.
\newblock Skills development for software engineers: Systematic literature review.
\newblock {\em Information and Software Technology}, 168:107395, 2024.

\bibitem{Idrus2018}
Hairuzila Idrus and Muhammad Ridhuan Tony~Lim Abdullah.
\newblock Implementation of pbl to enhance the soft skills of engineering students.
\newblock {\em SHS Web of Conferences}, 53:03008, 2018.

\bibitem{Zheng2015}
Guangzhi Zheng, Chi Zhang, and Lei Li.
\newblock Practicing and evaluating soft skills in it capstone projects.
\newblock In {\em Proceedings of the 16th Annual Conference on Information Technology Education}, SIGITE '15, pages 109--113, New York, NY, USA, 2015. Association for Computing Machinery.

\bibitem{Lee2024}
En-Shiun~Annie Lee, Luki Danukarjanto, Sadia Sharmin, Shou-Yi Hung, Sicong Huang, and Tong Su.
\newblock Exploring student motivation in integration of soft skills training within three levels of computer science programs.
\newblock In {\em Proceedings of the 55th ACM Technical Symposium on Computer Science Education V. 1}, SIGCSE 2024, pages 708--714, New York, NY, USA, 2024. Association for Computing Machinery.

\bibitem{Cukierman2014}
Uriel~Rubén Cukierman and Juan~María Palmieri.
\newblock Soft skills in engineering education: A practical experience in an undergraduate course.
\newblock In {\em 2014 International Conference on Interactive Collaborative Learning (ICL)}, pages 237--242, 2014.

\bibitem{Naiem2015}
Sarah Naiem, Mahmoud Abdellatif, and Salama E.
\newblock Evaluation of computer science and software engineering undergraduate’s soft skills in egypt from student’s perspective.
\newblock {\em Computer and Information Science}, 8:36--53, 2015.

\bibitem{Steele2023}
Jennifer~L. Steele.
\newblock To gpt or not gpt? empowering our students to learn with ai.
\newblock {\em Computers and Education: Artificial Intelligence}, 5:100160, 2023.

\bibitem{Huber2024}
Stefan~E Huber, Kristian Kiili, Steve Nebel, Richard~M Ryan, Michael Sailer, and Manuel Ninaus.
\newblock Leveraging the potential of large language models in education through playful and game-based learning.
\newblock {\em Educational Psychology Review}, 36:25, 2024.

\bibitem{Fleckenstein2024}
Johanna Fleckenstein, Jennifer Meyer, Thorben Jansen, Stefan~D Keller, Olaf Köller, and Jens Möller.
\newblock Do teachers spot ai? evaluating the detectability of ai-generated texts among student essays.
\newblock {\em Computers and Education: Artificial Intelligence}, 6:100209, 2024.

\bibitem{Brin2023}
Dana Brin, Vera Sorin, Akhil Vaid, Ali Soroush, Benjamin~S Glicksberg, Alexander~W Charney, Girish Nadkarni, and Eyal Klang.
\newblock Comparing chatgpt and gpt-4 performance in usmle soft skill assessments.
\newblock {\em Scientific Reports}, 13:16492, 2023.

\bibitem{Ali2023}
Sajid Ali, Tamer Abuhmed, Shaker El-Sappagh, Khan Muhammad, Jose~M Alonso-Moral, Roberto Confalonieri, Riccardo Guidotti, Javier Del~Ser, Natalia Díaz-Rodríguez, and Francisco Herrera.
\newblock Explainable artificial intelligence (xai): What we know and what is left to attain trustworthy artificial intelligence.
\newblock {\em Information Fusion}, 99:101805, 2023.

\bibitem{Vermesan2023}
Ovidiu Vermesan, Vincenzo Piuri, Fabio Scotti, Angelo Genovese, Ruggero~Donida Labati, and Pasquale Coscia.
\newblock Explainability and interpretability concepts for edge ai systems.
\newblock In {\em Advancing Edge Artificial Intelligence: System Contexts}, pages 197--227. Taylor \& Francis, New York, 2023.

\bibitem{Naamati2024}
Lior Naamati-Schneider.
\newblock Enhancing ai competence in health management: students’ experiences with chatgpt as a learning tool.
\newblock {\em BMC Medical Education}, 24:598, 2024.

\bibitem{Moulin2024}
Thiago~C Moulin.
\newblock Learning with ai language models: Guidelines for the development and scoring of medical questions for higher education.
\newblock {\em Journal of Medical Systems}, 48:45, 2024.

\bibitem{deFine2023}
Karl De~Fine~Licht.
\newblock Integrating large language models into higher education: Guidelines for effective implementation.
\newblock {\em Computer Sciences \& Mathematics Forum}, 8(1):65, 2023.

\bibitem{Hung2016}
Shao-Ting~Alan Hung.
\newblock Enhancing feedback provision through multimodal video technology.
\newblock {\em Computers \& Education}, 98:90--101, 2016.

\bibitem{Worsley2018}
Marcelo Worsley and Paulo Blikstein.
\newblock A multimodal analysis of making.
\newblock {\em International Journal of Artificial Intelligence in Education}, 28:385--419, 2018.

\bibitem{Fjortoft2020}
Henning Fjørtoft.
\newblock Multimodal digital classroom assessments.
\newblock {\em Computers \& Education}, 152:103892, 2020.

\bibitem{Du2023}
Xu~Du, Miao Dai, Hengtao Tang, Jui-Long Hung, Hao Li, and Jinqiu Zheng.
\newblock A multimodal analysis of college students’ collaborative problem solving in virtual experimentation activities: a perspective of cognitive load.
\newblock {\em Journal of Computing in Higher Education}, 35:272--295, 2023.

\bibitem{Oh2024}
Changdae Oh, Minhoi Park, Sungjun Lim, and Kyungwoo Song.
\newblock Language model-guided student performance prediction with multimodal auxiliary information.
\newblock {\em Expert Systems with Applications}, 250:123960, 2024.

\bibitem{Walkington2023}
Candace Walkington, Mitchell~J Nathan, Wen Huang, Jonathan Hunnicutt, and Julianna Washington.
\newblock Multimodal analysis of interaction data from embodied education technologies.
\newblock {\em Educational technology research and development}, 72:2565--2584, 2023.

\bibitem{Petkovic2025}
Uro{\v{s}} Petkovi{\'{c}}, Jonas Frenkel, Olaf Hellwich, and Rebecca Lazarides.
\newblock Nonverbal immediacy analysis in education: A multimodal computational model.
\newblock In Oliver Brock and Jeffrey Krichmar, editors, {\em From Animals to Animats 17}, pages 326--338, Cham, 2025. Springer Nature Switzerland.

\bibitem{deNovais2024}
André~Seixas De~Novais, José~Alexandre Matelli, and Messias~Borges Silva.
\newblock Fuzzy soft skills assessment through active learning sessions.
\newblock {\em International Journal of Artificial Intelligence in Education}, 34:416--451, 2024.

\bibitem{zadeh2002granular}
Lotfi~A Zadeh.
\newblock Granular computing as a basis for a computational theory of perceptions.
\newblock In {\em 2002 IEEE World Congress on Computational Intelligence. 2002 IEEE International Conference on Fuzzy Systems. FUZZ-IEEE'02. Proceedings (Cat. No. 02CH37291)}, volume~1, pages 564--565. IEEE, 2002.

\bibitem{TRIVINO201322}
Gracian Trivino and Michio Sugeno.
\newblock Towards linguistic descriptions of phenomena.
\newblock {\em International Journal of Approximate Reasoning}, 54(1):22--34, 2013.

\bibitem{CONDECLEMENTE201746}
Patricia Conde-Clemente, Jose~M. Alonso, Éldman O.~Nunes, Angel Sanchez, and Gracian Trivino.
\newblock New types of computational perceptions: Linguistic descriptions in deforestation analysis.
\newblock {\em Expert Systems with Applications}, 85:46--60, 2017.

\bibitem{de2022natural}
Andrea De~Anda-Trasvi{\~n}a, Alejandra Nieto-Garibay, and Joaqu{\'\i}n Guti{\'e}rrez.
\newblock Natural language report of the composting process status using linguistic perception.
\newblock {\em Applied Soft Computing}, 127:109357, 2022.

\bibitem{Li2024}
Dandan Li, Xiaolei Fan, and Lingchao Meng.
\newblock Development and validation of a higher-order thinking skills (hots) scale for major students in the interior design discipline for blended learning.
\newblock {\em Scientific Reports}, 14:20287, 2024.

\bibitem{Beagon2023}
Una Beagon, Klara Kövesi, Brad Tabas, Bente Nørgaard, Riitta Lehtinen, Brian Bowe, Christiane Gillet, and Claus~Monrad Spliid.
\newblock Preparing engineering students for the challenges of the sdgs: what competences are required?
\newblock {\em European Journal of Engineering Education}, 48(1):1--23, 2023.

\bibitem{Thornhill2023}
Branden Thornhill-Miller, Anaëlle Camarda, Maxence Mercier, Jean-Marie Burkhardt, Tiffany Morisseau, Samira Bourgeois-Bougrine, Florent Vinchon, Stephanie El~Hayek, Myriam Augereau-Landais, Florence Mourey, Cyrille Feybesse, Daniel Sundquist, and Todd Lubart.
\newblock Creativity, critical thinking, communication, and collaboration: Assessment, certification, and promotion of 21st century skills for the future of work and education.
\newblock {\em Journal of Intelligence}, 11(3):54, 2023.

\bibitem{Wang2020}
Chongyang Wang, Min Peng, Tao Bi, and Tong Chen.
\newblock Micro-attention for micro-expression recognition.
\newblock {\em Neurocomputing}, 410:354--362, 2020.

\bibitem{Li2021}
Yante Li, Xiaohua Huang, and Guoying Zhao.
\newblock Micro-expression action unit detection with spatial and channel attention.
\newblock {\em Neurocomputing}, 436:221--231, 2021.

\bibitem{Mencar2019}
Corrado Mencar and Jos{\'e}~M. Alonso.
\newblock Paving the way to explainable artificial intelligence with fuzzy modeling.
\newblock In Robert Full{\'e}r, Silvio Giove, and Francesco Masulli, editors, {\em Fuzzy Logic and Applications}, pages 215--227, Cham, 2019. Springer International Publishing.

\bibitem{Newmarch2017}
Jan Newmarch.
\newblock Ffmpeg/libav.
\newblock In {\em Linux Sound Programming}, chapter~12, pages 227--234. Apress, Berkeley, CA, 2017.

\bibitem{Jadoul2018}
Yannick Jadoul, Bill Thompson, and Bart {De Boer}.
\newblock Introducing parselmouth: A python interface to praat.
\newblock {\em Journal of Phonetics}, 71:1--15, 2018.

\bibitem{Radford2023}
A~Radford, J~W Kim, T~Xu, G~Brockman, C~McLeavey, and I~Sutskever.
\newblock Robust speech recognition via large-scale weak supervision.
\newblock In A~Krause, E~Brunskill, K~Cho, B~Engelhardt, S~Sabato, and J~Scarlett, editors, {\em Proceedings of the 40th International Conference on Machine Learning}, volume 202, pages 28492--28518, Honolulu, 2023. ML Research Press.

\bibitem{vasiliev2020}
Yuli Vasiliev.
\newblock {\em Natural Language Processing with Python and SpaCy: A Practical Introduction}.
\newblock No Starch Press, San Francisco, 2020.

\bibitem{Kumari2023}
Ainampudi Kumari~Sirivarshitha, Kadavakollu Sravani, Kothamasu~Santhi Priya, and Vasantha Bhavani.
\newblock An approach for face detection and face recognition using opencv and face recognition libraries in python.
\newblock In {\em 2023 9th International Conference on Advanced Computing and Communication Systems (ICACCS)}, volume~1, pages 1274--1278, 2023.

\bibitem{Viola2001}
P~Viola and M~Jones.
\newblock Rapid object detection using a boosted cascade of simple features.
\newblock In {\em Proceedings of the 2001 IEEE Computer Society Conference on Computer Vision and Pattern Recognition. CVPR 2001}, volume~1, pages I--I, 2001.

\bibitem{Taigman2014}
Yaniv Taigman, Ming Yang, Marc'Aurelio Ranzato, and Lior Wolf.
\newblock Deepface: Closing the gap to human-level performance in face verification.
\newblock In {\em 2014 IEEE Conference on Computer Vision and Pattern Recognition}, pages 1701--1708, 2014.

\bibitem{Chandel2023}
Ritika Chandel, Rajnandini Bhowmick, and U.~Hariharan.
\newblock A comparison of face landmark detection techniques.
\newblock In {\em 2023 4th International Conference on Computation, Automation and Knowledge Management (ICCAKM)}, pages 1--6, 2023.

\bibitem{Touvron2023}
Hugo Touvron, Louis Martin, Kevin Stone, Peter Albert, Amjad Almahairi, Yasmine Babaei, Nikolay Bashlykov, Soumya Batra, Prajjwal Bhargava, Shruti Bhosale, Dan Bikel, Lukas Blecher, Cristian~Canton Ferrer, Moya Chen, Guillem Cucurull, David Esiobu, Jude Fernandes, Jeremy Fu, Wenyin Fu, Brian Fuller, Cynthia Gao, Vedanuj Goswami, Naman Goyal, Anthony Hartshorn, Saghar Hosseini, Rui Hou, Hakan Inan, Marcin Kardas, Viktor Kerkez, Madian Khabsa, Isabel Kloumann, Artem Korenev, Punit~Singh Koura, Marie-Anne Lachaux, Thibaut Lavril, Jenya Lee, Diana Liskovich, Yinghai Lu, Yuning Mao, Xavier Martinet, Todor Mihaylov, Pushkar Mishra, Igor Molybog, Yixin Nie, Andrew Poulton, Jeremy Reizenstein, Rashi Rungta, Kalyan Saladi, Alan Schelten, Ruan Silva, Eric~Michael Smith, Ranjan Subramanian, Xiaoqing~Ellen Tan, Binh Tang, Ross Taylor, Adina Williams, Jian~Xiang Kuan, Puxin Xu, Zheng Yan, Iliyan Zarov, Yuchen Zhang, Angela Fan, Melanie Kambadur, Sharan Narang, Aurelien Rodriguez, Robert Stojnic, Sergey Edunov, and Thomas
  Scialom.
\newblock Llama 2: Open foundation and fine-tuned chat models, 2023.

\bibitem{Bhattacharya2024}
Ranjeeta Bhattacharya.
\newblock Strategies to mitigate hallucinations in large language models.
\newblock {\em Applied Marketing Analytics}, 10(1):62--67, 2024.

\bibitem{roman2023evaluating}
Daniel Rom{\'a}n-S{\'a}nchez, Jos{\'e}~Manuel De-La-Fuente-Rodr{\'\i}guez, Alberto Paramio, Juan~Carlos Paramio-Cuevas, Isabel Lepiani-D{\'\i}az, and Mar{\'\i}a-Reyes L{\'o}pez-Millan.
\newblock Evaluating satisfaction with teaching innovation, its relationship to academic performance and the application of a video-based microlearning.
\newblock {\em Nursing Open}, 10(9):6067--6077, 2023.

\end{thebibliography}
\newpage

\begin{appendices}

\section{Raw Student Ratings for Correlation Analysis}\label{secA0}

\begin{table}[htp!]
\caption{Individual Ratings per Task Used for the Correlation Analysis}
    \label{tab_grades}
    \centering
    \scriptsize
\begin{tabular}{|c|c|c|c|c|}
\hline
 \textbf{Group} & \textbf{Student} & \textbf{Task 1} & \textbf{Task 2} & \textbf{Task 3} \\ \hline
\multirow{10}{*}{Machine learning 2022} & A1      & 85     & -      & -      \\
                       & A2      & 83     & -      & -      \\
                       & A3      & 86     & -      & -      \\
                       & A4      & 83     & -      & -      \\
                       & A5      & 87     & -      & -      \\
                       & A6      & 82     & -      & -      \\
                       & A7      & 86     & -      & -      \\
                       & A8      & 85     & -      & -      \\
                       & A9      & 85     & -      & -      \\
                       & A10     & 88     & -      & -      \\ \hline
\multirow{18}{*}{Machine learning 2023}  & B1      & 88     & 82     & 87     \\
                       & B2      & 86     & 84     & 87     \\
                       & B3      & 82     & 83     & 87     \\
                       & B4      & 82     & 82     & 83     \\
                       & B5      & 88     & 86     & 88     \\
                       & B6      & 87     & 85     & 84     \\
                       & B7      & 86     & 82     & 83     \\
                       & B8      & 82     & 85     & 83     \\
                       & B9      & 87     & 85     & 84     \\
                       & B10     & 82     & 83     & 82     \\
                       & B11     & 83     & 85     & 82     \\
                       & B12     & 84     & 82     & 82     \\
                       & B13     & 85     & 86     & 84     \\
                       & B14     & 85     & 85     & 84     \\
                       & B15     & 89     & 83     & 89     \\
                       & B16     & 85     & 84     & 86     \\
                       & B17     & 86     & 85     & 83     \\
                       & B18     & 85     & 83     & 84     \\ \hline
\multirow{21}{*}{Interaction Human-Computer}  & CA1     & 88     & 87     & 82     \\
                       & CA2     & 95     & 86     & 100    \\
                       & CA3     & 86     & 82     & 86     \\
                       & CA4     & 95     & 84     & 90     \\
                       & CA5     & 88     & 85     & 82     \\
                       & CA6     & 94     & 98     & 84     \\
                       & CA7     & 97     & 89     & 85     \\
                       & CA8     & 97     & 84     & 92     \\
                       & CA9     & 94     & 82     & 82     \\
                       & CA10    & 84     & 87     & 88     \\
                       & CB1     & 92     & -      & 84     \\
                       & CB2     & 91     & -      & 95     \\
                       & CB3     & 88     & -      & 83     \\
                       & CB4     & 88     & -      & 86     \\
                       & CB5     & 93     & -      & 92     \\
                       & CB6     & 96     & -      & 87     \\
                       & CB7     & 92     & -      & 85     \\
                       & CB8     & 93     & -      & 82     \\
                       & CB9     & 98     & -      & 93     \\
                       & CB10    & 99     & -      & 85     \\
                       & CB11    & 95     & -      & 87    \\ \hline
\end{tabular}
\end{table}

\newpage

\section{Detailed Measures for Teacher Satisfaction Evaluation}\label{secA1}

\begin{table}[!htp]
    \caption{Measures for Teacher Satisfaction Evaluation}
    \label{tab_satisfaction_evaluation_explained}
    \centering
    \small
    \begin{tabular}{|p{4cm}|p{7.8cm}|}
     \hline
     \textbf{Measure} & \textbf{Explanation} \\
     \hline
      Transparency and Consistency & The methodology employed to assess soft skills is transparent, consistent, and technically sound, relying on valid and reliable data, observations, and interpretations \\ \hline

     Impartial Assessment & The assessment of soft skills is conducted impartially, thereby ensuring that all students are treated equally and objectively \\ \hline

     Feedback Support & Learners are supported in understanding and responding to the feedback they receive, helping them develop a positive mindset and Feedback Orientation (FO) \\ \hline

     Accountability and Reflection & The assessment of soft skills facilitates the development of accountability and encourages a reflective approach to personal experiences \\ \hline

     Performance Alignment & Following the completion of the soft skills assessment, structured opportunities are made available to facilitate the alignment of current abilities with the desired level of performance \\ \hline

     Motivation and Feedback Value & The evaluation of soft skills has been demonstrated to enhance motivation and reinforce the value of actively seeking and receiving feedback \\ \hline

     Instructional Insights & Soft skills assessment provides educators with valuable insights that inform the development of future instructional and training strategies \\ \hline
    \end{tabular}
\end{table}

\section{Student Satisfaction and Feedback Question Mapping}
\begin{table}[!htp]
    \caption{Related Questions for Students Satisfaction and Feedback}
    \label{tab_satisfaction_feedback_explained}
    \centering
    \small
    \begin{tabular}{|p{4cm}|p{8cm}|}
     \hline
     \textbf{Aspect to Evaluate} & \textbf{Related Question} \\ \hline

     Understanding of Subject & The tasks help me to understand the subject. \\ \hline
     
     Exam Preparation & The tasks help me prepare for the theory exam's objective test. \\ \hline
     
     Interest in Subject & The tasks help me to increase my interest in the subject. \\ \hline
     
     Participation Motivation & The tasks help me to motivate my participation in the subject \\ \hline
     
     Didactic Resource Quality & The tasks are a good didactic resource. \\ \hline
     
     Task Timing Satisfaction & The proposed time for the start of the tasks is satisfactory \\ \hline
     
     Task Completion Time & The time provided for the implementation of the tasks is sufficient. \\ \hline
     
     Satisfaction from Rewards & The rewards obtained through overcoming the teaching innovation have increased your satisfaction towards the study of the subject \\ 
     \hline
    \end{tabular}
\end{table}

\end{appendices}

\end{document}